\documentclass[conference]{IEEEtran}

\usepackage{fancyhdr}
\usepackage[normalem]{ulem}
\usepackage[hyphens]{url}
\usepackage{microtype}
\usepackage{support-caption}
\usepackage{subcaption}
\usepackage{adjustbox}
\usepackage{tabu}
\usepackage{multirow}
\usepackage{comment}
\usepackage{cite}
\usepackage[fleqn]{amsmath}
\usepackage{amssymb,amsfonts}
\usepackage{algorithmic}
\usepackage{graphicx}
\usepackage{textcomp}
\usepackage{xcolor}
\usepackage{fancyhdr}
\usepackage[hyphens]{url}
\usepackage{nicefrac}
\definecolor{fontblue}{RGB}{22, 0, 186}

\usepackage{eso-pic}

\usepackage[bookmarks=true,breaklinks=true,colorlinks,linkcolor=black,citecolor=blue,urlcolor=black]{hyperref}

\def\BibTeX{{\rm B\kern-.05em{\sc i\kern-.025em b}\kern-.08em
    T\kern-.1667em\lower.7ex\hbox{E}\kern-.125emX}}

\newcommand{\eg}{\emph{e.g.,}}
\newcommand{\ie}{\emph{i.e.,}}

\title{Term Revealing: Furthering Quantization at Run Time on Quantized DNNs} 
\author{
\IEEEauthorblockN{H. T. Kung\textsuperscript{*}}
\IEEEauthorblockA{\textit{Harvard University} \\
kung@harvard.edu}
\and
\IEEEauthorblockN{Bradley McDanel\textsuperscript{*}}
\IEEEauthorblockA{\textit{Harvard University} \\
mcdanel@fas.harvard.edu}
\and
\IEEEauthorblockN{Sai Qian Zhang\textsuperscript{*}}
\IEEEauthorblockA{\textit{Harvard University} \\
zhangs@g.harvard.edu}
}

\begin{document}
\maketitle
\begingroup\renewcommand\thefootnote{*}
\footnotetext{Equal contribution ordered alphabetically.}
\endgroup
\pagestyle{plain}

\AddToShipoutPictureBG*{%
  \AtPageUpperLeft{%
    \hspace{210mm}%
    \raisebox{-1\baselineskip}{%
      \makebox[0pt][r]{This preliminary version has been accepted by}}
    \raisebox{-2\baselineskip}{%
      \makebox[0pt][r]{Intl. Conf. for High Performance Computing,}}
    \raisebox{-3\baselineskip}{%
      \makebox[0pt][r]{Networking, Storage, and Analysis (SC20).}}    
      }}%

\begin{abstract}
We present a novel technique, called Term Revealing (TR), for \textit{furthering quantization} at run time for improved performance of Deep Neural Networks (DNNs) already quantized with conventional quantization methods. TR operates on power-of-two terms in binary expressions of values. In computing a dot-product computation, TR dynamically selects a fixed number of largest terms to use from the values of the two vectors in the dot product. By exploiting normal-like weight and data distributions typically present in DNNs, TR has a minimal impact on DNN model performance (i.e., accuracy or perplexity). We use TR to facilitate tightly synchronized processor arrays, such as systolic arrays, for efficient parallel processing. We show an FPGA implementation that can use a small number of control bits to switch between conventional quantization and TR-enabled quantization with a negligible delay. To enhance TR’s efficiency further, we use a signed digit representation (SDR), as opposed to classic binary encoding with only nonnegative power-of-two terms. To perform conversion from binary to SDR, we develop an efficient encoding method called HESE (Hybrid Encoding for Signed Expressions) that can be performed in one pass looking at only two bits at a time. We evaluate TR with HESE encoded values on an MLP for MNIST, multiple CNNs for ImageNet, and an LSTM for Wikitext-2, and show significant reductions in inference computations (between 3-10x) compared to conventional quantization for the same level of model performance.
\end{abstract}

\section{Introduction}
Deep Neural Networks (DNNs) have achieved state-of-the-art performance across a variety of domains, including Recurrent Neural Networks (RNNs) and Transformers for natural language processing and Convolutional Neural Networks (CNNs) for computer vision. However, the high computation complexity of DNNs makes them expensive to deploy at scale in datacenter contexts, as a popular model (e.g., Google’s Smart Compose email autocomplete~\cite{smartcompose}) may be queried millions of times per day, with each query requiring 10s to 100s of GFLOPs.

To address these high computational costs, significant research effort has been spent on developing techniques that reduce the computational complexity of pre-trained DNNs. One of the most commonly used techniques is post-training quantization~(see, e.g., \cite{lin2016fixed}), where 32-bit floating-point DNN weights and data (activations) are converted to a fixed-point representation (e.g., 8-bit fixed-point) to reduce the amount of computation performed per inference sample. One benefit of post-training quantization is that it does not require access to the original training dataset, and can therefore be applied by a third-party (such as a cloud service) as a step to reduce costs. In this work, we define a \textit{term} as a nonzero bit in a quantized fixed-point value. For instance, we say that the 8-bit value 5 (00000101) is composed of two terms: $2^2 + 2^0$.

\begin{figure}
    \centering
    \includegraphics[width=\columnwidth]{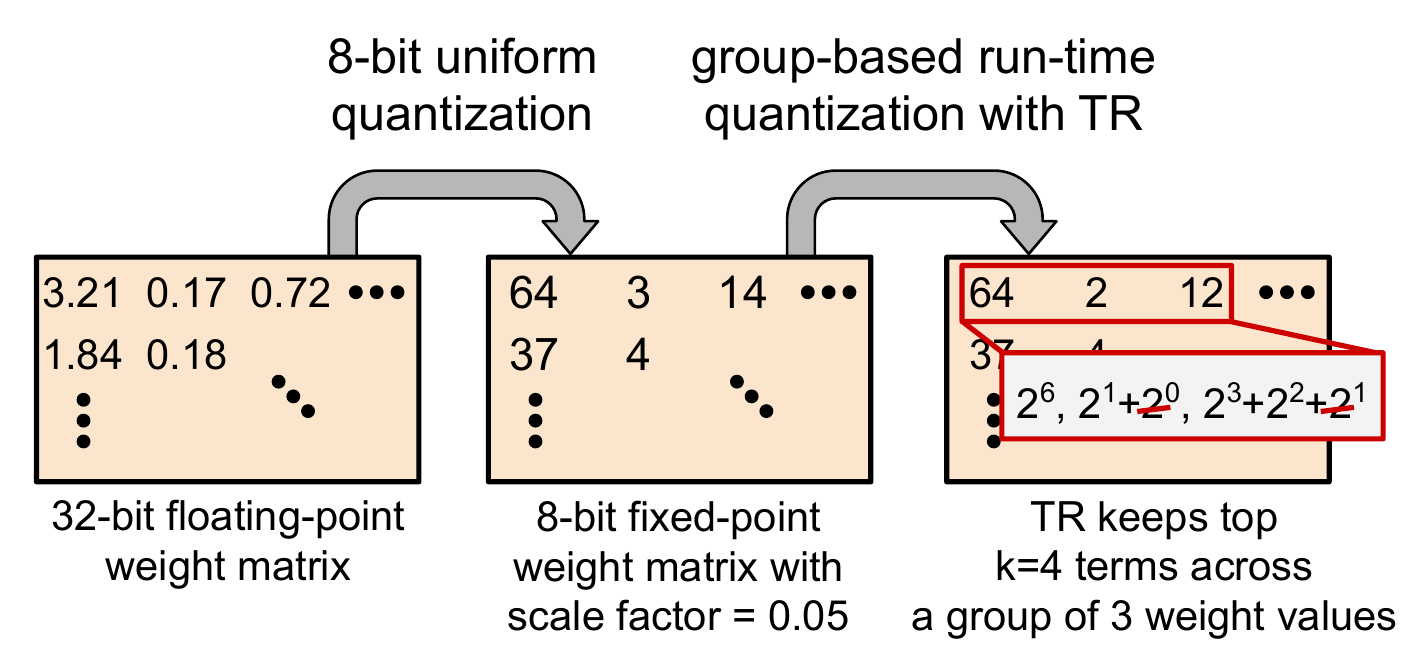}
    \caption{In conventional quantization, a 32-bit floating point weight matrix (left) is converted to an 8-bit fixed-point format via uniform quantization (middle). We propose to further quantization via Term Revealing (TR) which is a group-based run-time quantization method (right). By limiting the number of power-of-two terms across a group of values, TR enables a tighter processing bound for DNN dot product computations.}
    \label{fig:high-level}
\end{figure}

In this paper, we \textit{further quantize} the computation of an already quantized DNN at run time to realize additional computation savings. That is, we propose to perform further run-time quantization on, for example, a quantized 8-bit DNN while still achieving the same level of model performance. Note that for furthering quantization, we must use new techniques beyond conventional quantization methods, for otherwise, the original DNN could have been quantized to a lower precision in the first place (e.g., a 4-bit DNN instead of an 8-bit DNN).

Specifically, we introduce a novel group-based quantization method, which we call \textit{Term Revealing} (TR). TR, shown in Figure~\ref{fig:high-level}, ranks the terms in a \textit{group} of weight values associated with a dot-product computation to reveal a fixed number of top terms (called a \textit{group budget}) to use for the dot-product computation. By limiting the number of terms to a group budget $k$ and pruning the remaining smaller terms, TR enables a more efficient implementation of dot-product computations in DNN. \textbf{With TR, the selected terms for a value are based on their relative rankings against terms of other values in the group.} TR's run-time group-based quantization, a departure from traditional individual value-based quantization, allows TR to carry out additional quantization on an already quantized DNN.

While allowing further quantization at run time, TR is able to achieve the same level of model performance as the original quantized DNNs for two reasons.  First, TR uses group-based term selection, which prunes only smaller terms in a group (e.g., $2^1$ and $2^0$ terms), leading to minimal added quantization error. For groups with fewer terms than the allocated group budget, no additional truncation needs to be performed. Second, by adapting to the dynamic range of current values in a group, TR with a small group budget can still discern differences in small weight and data values typically present in DNNs (see Section~\ref{sec:term-dist}), where these differences are critical in differentiating features.

With a simple FPGA design, we can use a small number of control bits to reconfigure a hardware supporting quantized computations under conventional quantization to one supporting run-time TR quantization, and vice versa.

To simplify our introduction of TR, we use conventional binary representations where all terms are nonnegative. However, shorter \textit{signed digit representations} (SDRs) which use both positive and negative terms, such as Booth encodings~\cite{booth1951signed}, can allow fewer terms in expressing a value and lead to increased computation savings in TR-enabled quantization. To this end, we have developed a new signed encoding called \textit{Hybrid Encoding for Shortened Expressions} (HESE), which produces SDRs with the theoretical minimum number of terms.

The novel contributions of the paper are:
\begin{itemize}
    \item The concept of run-time quantization on already quantized DNNs to realize further computation savings.
    \item A \textit{group-based term ranking mechanism}, called term revealing (TR) and its \textit{term MAC} (tMAC) hardware design for the implementation of our proposed further quantization at run time.
    \item An FPGA system which requires \textit{minimal reconfiguration} to efficiently supports both conventional quantization and TR-enabled quantization.                        
    \item A \textit{one-pass encoding method}, called Hybrid Encoding for Shortened Expressions (HESE), for converting conventional binary representations to minimum-length SDRs. Using fewer terms compared to previous signed representations, HESE enhances TR's computation efficiency. 
\end{itemize}

\section{Background and Related Work}
\label{sec:bg}

In Section~\ref{sec:other-designs}, we discuss related work on pruning and quantization techniques for performing efficient DNN inference. Then, in Section~\ref{sec:bg-bit-sparsity}, we discuss prior work on hardware architectures which aim to exploit bit-level sparsity. Finally, in Section~\ref{sec:matrix-computation} we illustrate how matrix multiplication is performed with systolic arrays. 

\subsection{Pruning and Quantization Methods}
\label{sec:other-designs}

There has been significant research efforts in pruning-based methods which exploit value-level sparsity in CNN weights, as performing multiplication with zero operands can be viewed as wasted computation~\cite{han2015deep,wen2016learning,huang2017condensenet,he2017channel,luo2017thinet,narang2017block,gray2017blocksparse,kung2019packing,ren2019admm}. However, these pruning methods typically require model retraining, making them not feasible for a third-party that is hosting the model (as it requires access to the full training dataset). Additionally, unstructured pruning methods which achieve the best performance (e.g.,~\cite{han2015deep}) are hard to implementing efficiently in special-purpose hardware, as the the remaining nonzero weights are randomly distributed. In this paper, we propose to further reduce the amount of computation even for nonzero values by exploiting bit-level sparsity as opposed to conventional value-level sparsity.

Quantization~\cite{courbariaux2014training,gupta2015deep,zhu2016trained,hubara2017quantized,park2017weighted,kapur2017low,zhou2017incremental,wang2017fixed,chen2018exploiting,park2018energy,park2018value,hu2018hashing,mcdanel19full,li2019bstc} lowers precision of the values in weights and data in order to reduce the associated storage, I/O and/or computation costs. However, aggressive post-training quantization (e.g., to 4-bit representations) introduces additional error into the computation, leading to decreased model performance. Due to this, many low-precision quantization approaches, such as binary neural networks~\cite{courbariaux2015binaryconnect}, must be performed during training. Our proposed TR approach is applied on top of 8-bit quantization and does not require additional training. Note that our approach does not reduce the precision of the weights (i.e., weights are still 8-bit fixed-point values after term revealing) but instead reduces the number of nonzero terms to be used at runtime across a group of weights.

\subsection{Hardware Architectures for Exploiting Bit-level Sparsity}
\label{sec:bg-bit-sparsity}
There has been growing interest in exploiting bit-level sparsity (i.e., the zero bits present in weight and data values) as opposed to value-level sparsity discussed in the previous section. A bit-level multiplication with a zero bit can be viewed as wasted computation in the same manner as a value-level multiplication with a zero value, in that both operations do not effect the result. Based on this observation, Bit-Pragmatic introduces an architecture that utilizes a nonzero term-based representation to remove multiplication with zero bits in weights while keeping data in a conventional representation~\cite{albericio2017bit}. Bit-Tactical follows up this work by grouping nonzero weight and data values to achieve more efficient scheduling of nonzero computation~\cite{delmas2018bit}. Both of these approaches assume 8-bit or 16-bit fixed-point quantization (i.e., the first step in Figure~\ref{fig:high-level}).

However, due to the more fine-grained nature of these bit-level architectures, efficiently scheduling bit-level operations across multiple groups of computations becomes challenging, as each group may have a different amount computation to perform. Generally, this leads to stragglers that require significantly more bit-level operations than other groups. Both Bit-Pragmatic and Bit-Tactical handle this straggler problem by adding a synchronization barrier which makes all groups wait until the straggler is finished. Due to this, in processing many groups concurrently, they can only exploit bit-level sparsity up to the degree of the group with most bit-level operations (i.e., the straggler). We find that this worse case can be a factor of 2-3$\times$ more bit-level operations compared to the average case. By comparison, TR provides a tighter processing bound which enables synchronous computation across all groups. This is done by removing smaller terms from groups with a large number of terms (the second quantization step in Figure~\ref{fig:high-level}).

\subsection{Systolic Arrays for Matrix Multiplication}
\label{sec:matrix-computation}

\begin{figure}
    \centering
    \includegraphics[width=0.85\columnwidth]{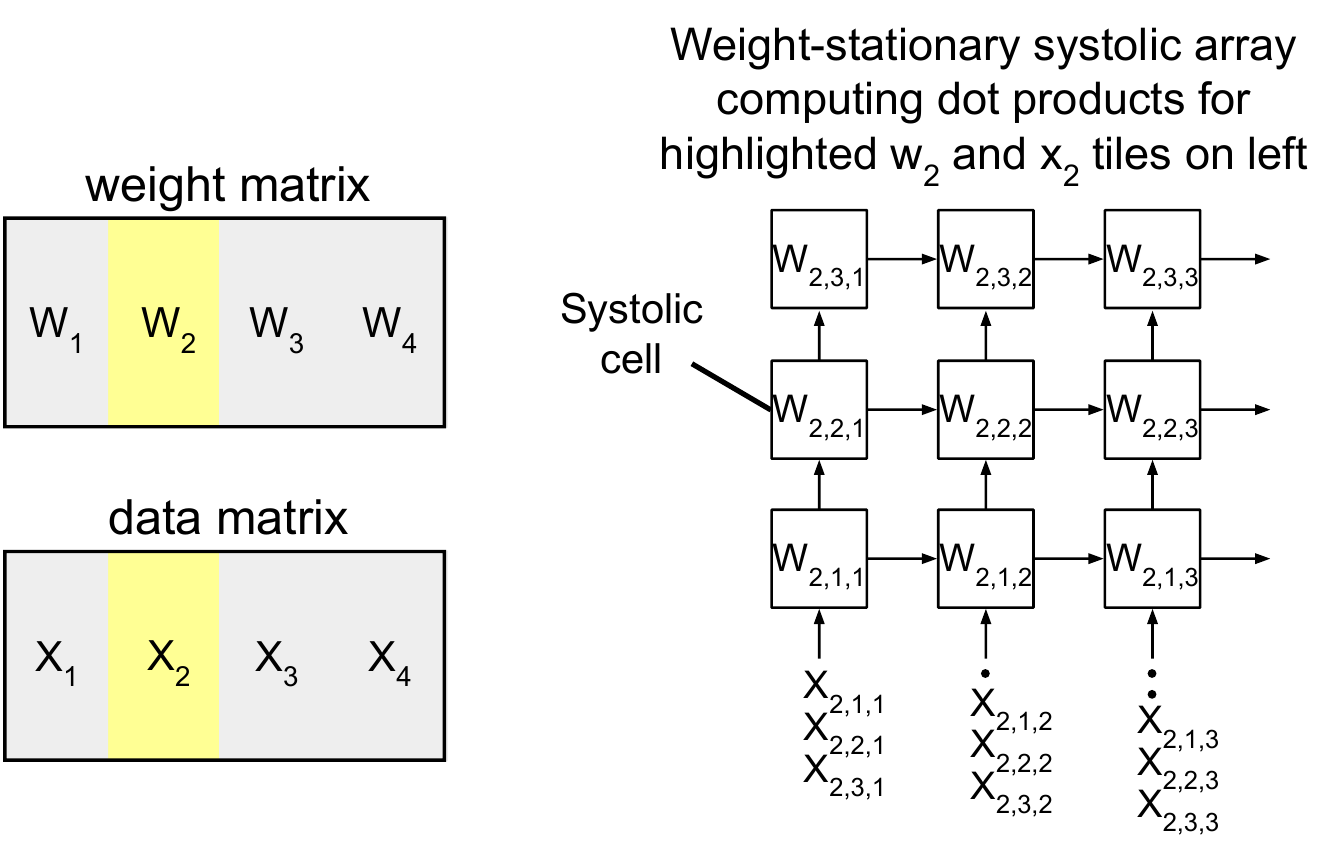}
    \caption{The weight matrix $W$ and data matrix $X$ for a layer in a DNN (left) are partitioned into four tiles to be processed in a systolic array (right). The highlighted weight tile ($W_2$) is shown loaded into the systolic array with the data tile ($X_2$) entering the systolic array from below.}
    \label{fig:systolic-array}
\end{figure}

The majority of computation in the forward propagation of a DNN consists of matrix multiplications between a learned weight matrix in each layer and input or data being propagated through the layer, as shown on the left side of Figure~\ref{fig:systolic-array}. Systolic arrays are known to be able to efficiently implement matrix multiplication due to their regular design, dataflow architectures and reduced memory access~\cite{kung1982systolic}. The right side of Figure~\ref{fig:systolic-array} shows a 3$\times$3 systolic array, for computing dot products between $W_2$ and $X_2$ (highlighted in the weight and data matrices). The data in the partition (\eg~$X_{2,1,1}$) are passed into the systolic array from below in a skewed fashion in order to maintain synchronization between cells. We use this systolic array design as the starting point for our FPGA system in Section~\ref{sec:term revealing-hw}.

\section{Term Revealing}
\label{sec:term revealing-approach}
In this section, we introduce a group-based quantization method called term revealing (TR), which is applied to quantized DNNs at run time.

\subsection{DNN Weight and Data Distributions}
\label{sec:term-dist}
\begin{figure}
    \centering
    \includegraphics[width=\columnwidth]{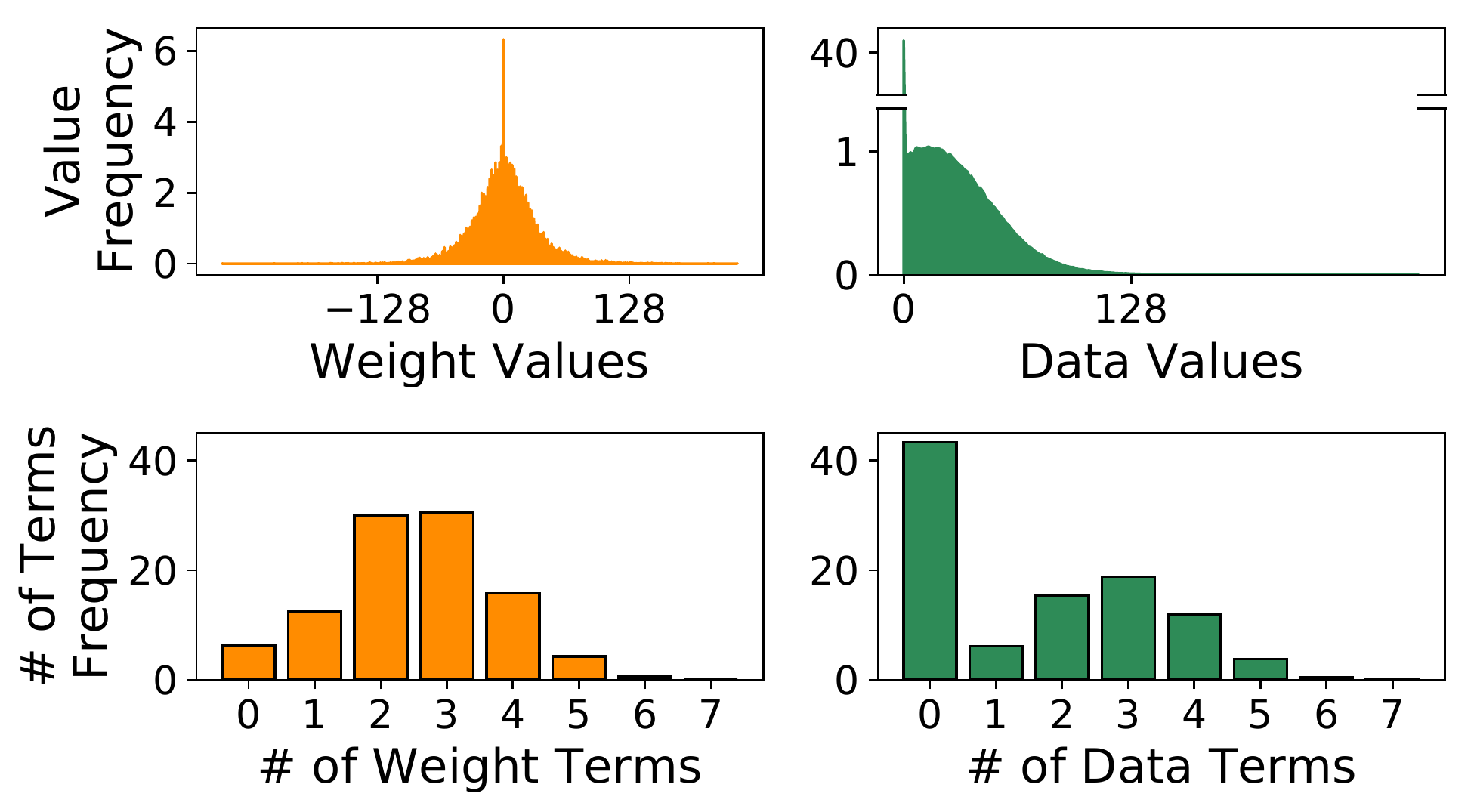}
    \caption{The distributions of weight and data values (top) shape the distribution of the number of terms in a binary encoding for both weights and data (bottom).}
    \label{fig:values-vs-terms-dist}
\end{figure}

As mentioned earlier, TR leverages weight and data distributions of DNNs.
DNNs are often trained with weight decay regularization to improve model generalization~\cite{zhang2016understanding} and batch normalization~\cite{ioffe2015batch} on data which both improves the stability of convergence and improves the performance of the learned model. A consequence is that the weights are approximately normally distributed and the data follow a half-normal distribution (as ReLU sets negative values to 0). Figure~\ref{fig:values-vs-terms-dist} (top row) illustrates these distributions for the weights in 7th convolution layer of ResNet-18~\cite{he2016deep} trained on ImageNet~\cite{deng2009imagenet} and the data input to the layer. Both the weights and data are quantized to 8-bit fixed-point using uniform quantization (QT). 

The higher frequency of small values means that most elements are represented with only 2 or 3 power-of-two terms as shown in Figure~\ref{fig:values-vs-terms-dist} (bottom). For instance, the value 6 is represented with two power-of-two terms $(2^2+2^1)$. In the figure, 79\% of weight values and 84\% of data are represented with 3 or fewer power-of-two terms. Note that the most significant bit (MSB) in the 8-bit representation is used to represent the sign of each value, thus each value has at most 7 terms.

\subsection{Computing Dot Products via Term Pair Multiplications}
\label{sec:term-pairs}

\begin{figure}
    \centering
    \includegraphics[width=\columnwidth]{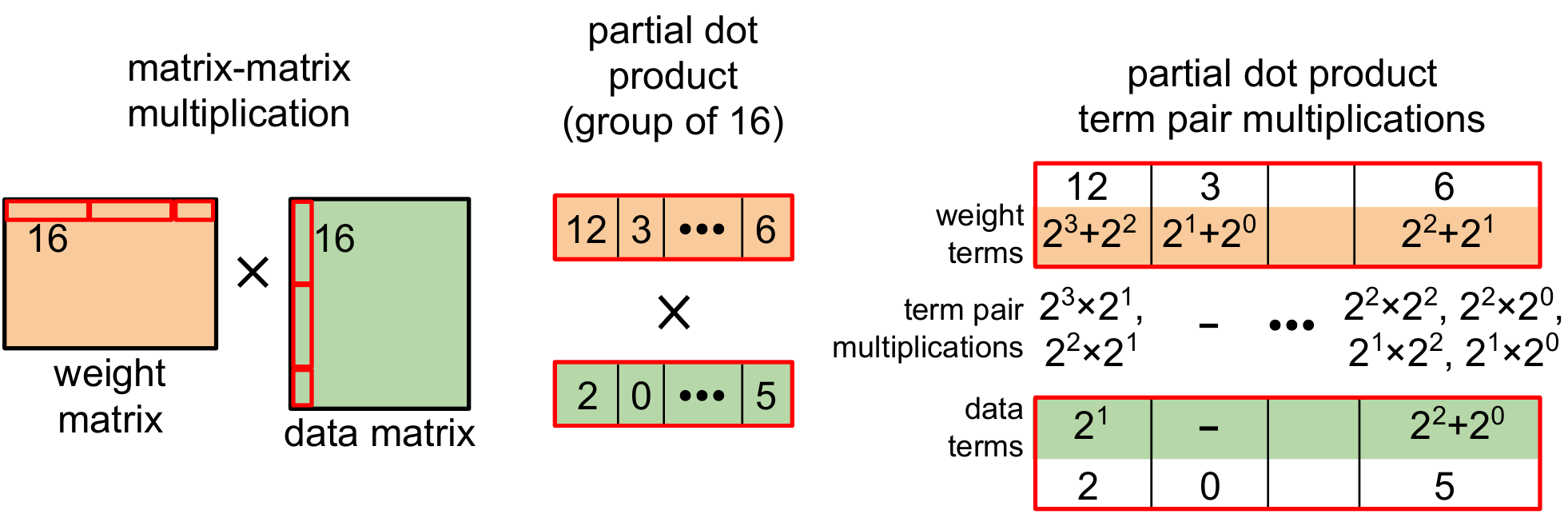}
    \caption{A matrix-matrix multiplication between a weight and data matrix (left) divided into partial dot products of length 16 (one partial dot product is shown in the middle). Each partial dot product is computed by multiplying all pairs of terms (right) across the 16 value in the weight and data vectors.}
    \label{fig:term-pair-mult}
\end{figure}

Assume that we compute dot products in matrix-matrix multiplication between quantized weights and data by dividing both vectors into groups of a given length (e.g., 16). This group-based formulation is motivated by efficient hardware implementations described later in Section~\ref{sec:term revealing-hw}. Figure~\ref{fig:term-pair-mult} illustrates how partial dot products, partitioned into groups of length 16, are computed using term pair multiplications. In the example, the first value in the weight vector $12 = 2^3 + 2^2$ multiplied with the first data value $2 = 2^1$ is computed using two term pair multiplications ($2^3\times 2^1 + 2^2\times2^1 = 2^{(3 + 1)} + 2^{(2 + 1)} = 2^4 + 2^3 = 24$) as shown on the right of Figure~\ref{fig:term-pair-mult}. Using this paradigm, we can analyze the number of term pair multiplications that are required per partial dot product (e.g., with a group size of 16) across all groups in a matrix-matrix multiplication.

\begin{figure}
  \centering
  \includegraphics[width=0.96\columnwidth]{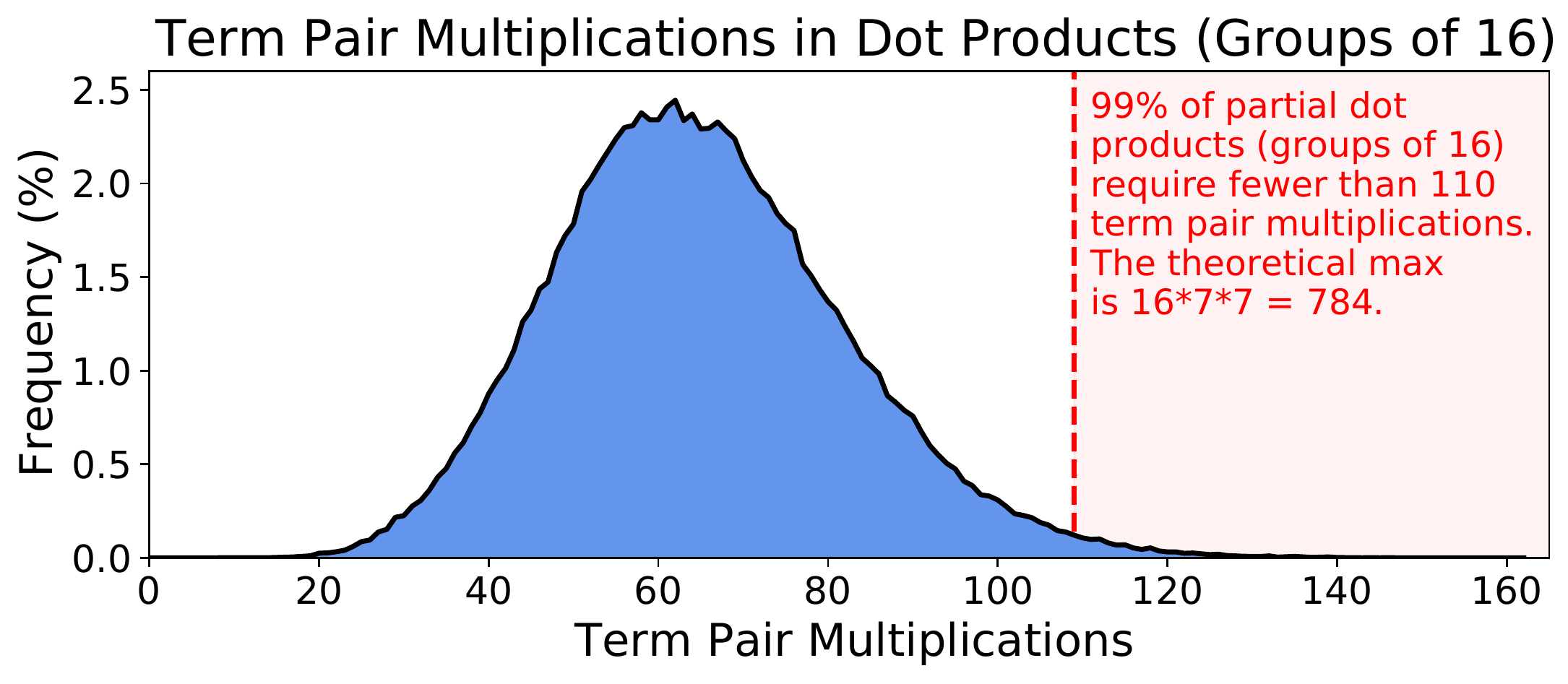}
\caption{The number of term pair multiplications required for partial dot products with groups of size 16 in an 8-bit DNN.}
\label{fig:term-group-dist}
\end{figure}

Figure~\ref{fig:term-group-dist} shows a histogram of the number of term pair multiplications for partial dot products with groups of 16 values in the 7th convolutional layer of ResNet-18. Interestingly, 99\% of these groups require under 110 term pair multiplications even though the theoretical maximum, where all weight and data values use 7 terms (i.e., every value is $127 = 2^6 + 2^5 + 2^4 + 2^3 + 2^2 + 2^1 + 2^0$), is $16\times 7 \times 7 = 784$. In this work, we propose to restrict the number of term pair multiplications performed in each partial dot product (e.g., to 110 instead of 784) in order to achieve tightly synchronized parallel processing across systolic cells. As we know that DNNs are robust in performance to small amounts of error (\eg~through the initial uniform quantization step), it is reasonable to expect that they would also tolerate an additional quantization step such as our proposed TR-enabled quantization that makes small modifications to enforce a tighter processing bound. We use term pairs multiplications as a proxy for the amount of computation performed during inferences, as the hardware system described in Section~\ref{sec:term revealing-hw} performs dot products using this term pair multiplication approach.

\subsection{Overview of Term Revealing}
\label{sec:tr-overview}

\begin{figure}
    \centering
    \includegraphics[width=0.9\columnwidth]{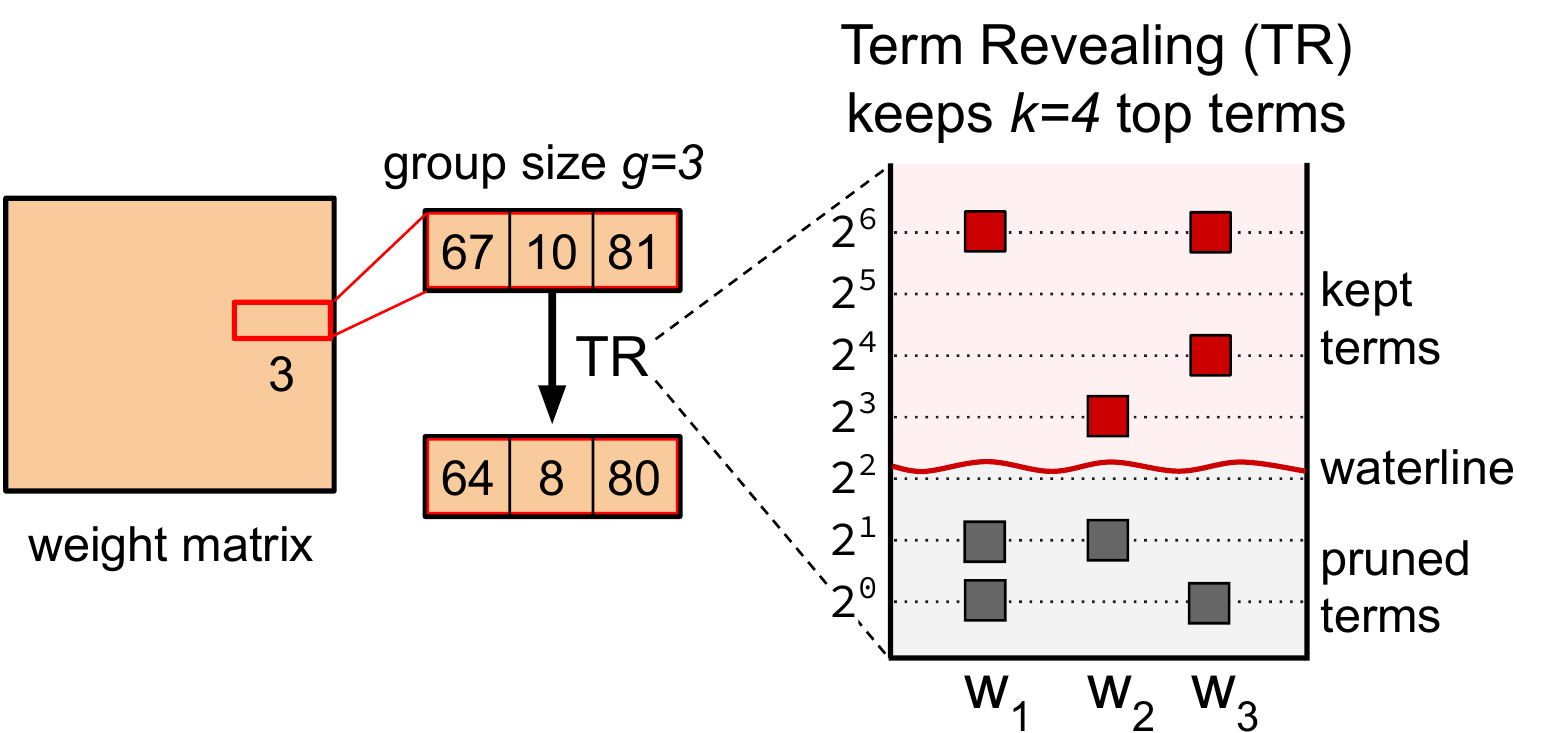}
    \caption{A weight matrix (left) is partitioned into groups of size 3. The elements of each group (middle) are passed into TR. The receding water algorithm (right) based on term ranking keeps the top $k=4$ terms (red) for the values in the group; the rest of the terms are pruned.}
    \label{fig:waterline}
\end{figure}

Term revealing is a group-based term ranking method that sets a limit on the number of terms allotted to a scheduling group. TR consists of three steps:

\begin{enumerate}
\item \textit{Grouping elements} as shown in Figure~\ref{fig:waterline} (left). For a given weight matrix, we partition it into equal size groups which are used in dot product computations. The \textit{group size} $g$, denotes the number of values per group and may assume various values such as 2, 3, 4, 8, 16, etc.

\item \textit{Configure a group budget $k$} which is used for every group. The budget bounds the number of terms used in dot-product computations across the values in a group.

\item \textit{Identify top $k$ terms} in the group using a \textit{receding water algorithm} that ranks and selects the terms as shown in Figure~\ref{fig:waterline} (right). It keeps the largest $k$ terms in a group and prunes the remaining smaller terms below a waterline. Note that some groups may have fewer than $k$ terms, meaning that no pruning occurs.
\end{enumerate}

Figure~\ref{fig:waterline} illustrate how TR is applied to a group of $g = 3$ values in a weight matrix with a term budget $k = 4$. The three values are decomposed into their term representations, and scanned row by row (viewed as a waterline), starting from the $2^6$ term and finishing at the $2^0$ term, until the group budget is reached. In the example, the group budget of $k=4$ is reached at the $2^3$ term for $w_2$. The remaining low-order terms (e.g., $2^2$ and below for this group) are pruned, adding a small amount of additional quantization error. For instance, after TR, $w_3$ is quantized from 81 to 80. 

\begin{figure}
    \centering
    \includegraphics[width=0.8\columnwidth]{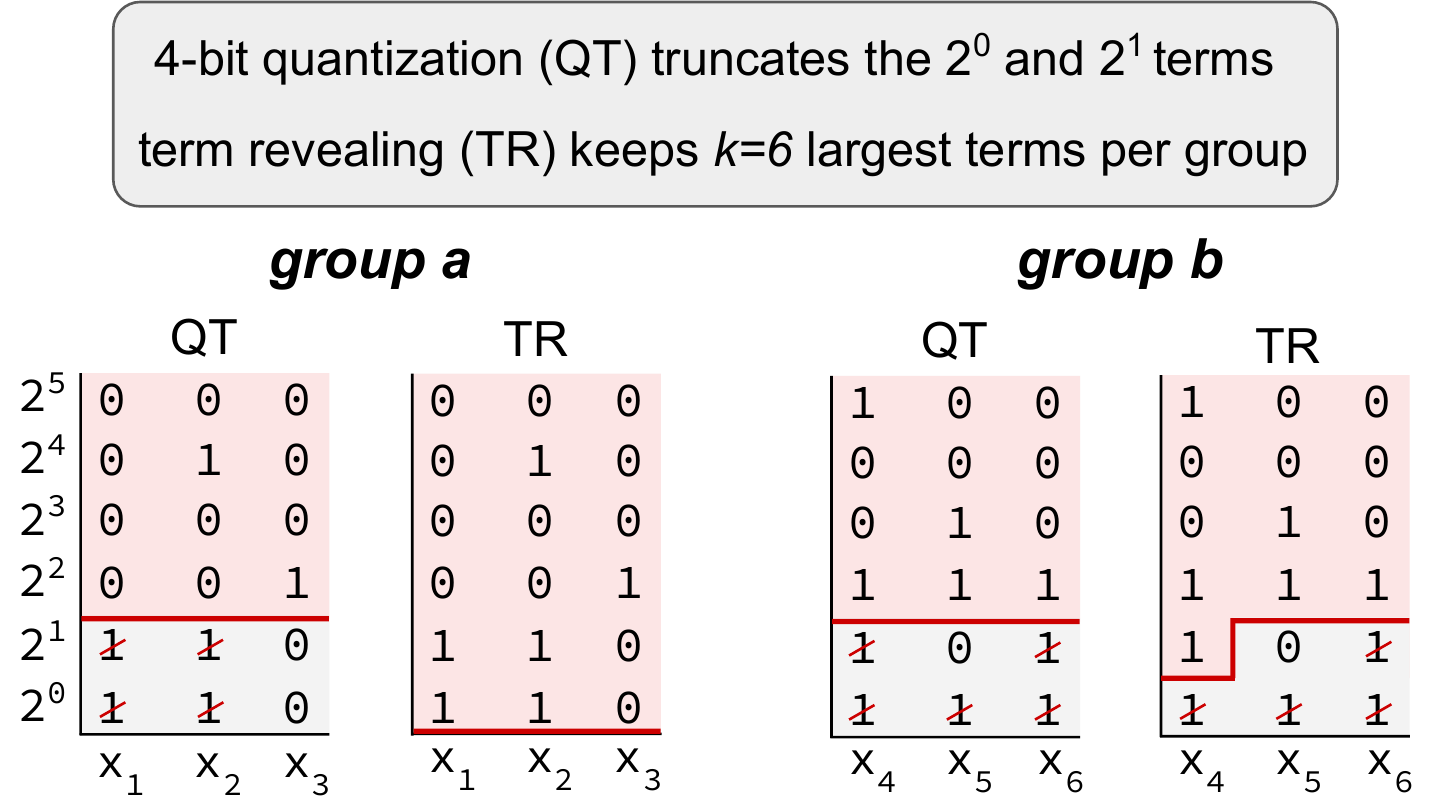}
    \caption{4-bit uniform quantization (QT) always truncates smaller terms (e.g., the $2^0$ and $2^1$ terms). This leads to large quantization error for groups with many small terms as in group a. In contrast, by keeping the top 6 terms, TR introduces significantly less quantization error on average. Additionally, TR reduces the number of term pair multiplications to $7 \times k = 42$, which is smaller than 4-bit QT of $7 \times 4 \times 3 = 84$ by a factor of $2\times$.}
    \label{fig:term-groups}
\end{figure}

Since the position of the waterline is determined by the distribution of terms in a group, the amount of pruning induced by TR varies across each group of values. Consider two groups of weights (group a: $w_1,w_2,w_3$), and (group b: $w_4,w_5,w_6$). Figure~\ref{fig:term-groups} illustrates the quantization error incurred when 4-bit QT (truncating the $2^0$ and $2^1$ terms) and TR (keeping the top $k=6$ terms) are applied to both groups. For group a, we see that TR introduces no error as the group has only 6 terms. By comparison, 4-bit QT introduces error by pruning all of the $2^0$ and $2^1$ terms, as conventional quantization keeps only the largest 4 terms across all values. For group b, which has significantly more terms, TR and 4-bit QT perform a similar amount of truncation. Group b represents a worse case for TR, as most groups will have significantly fewer terms. 

Therefore, in practice, we can use a small group budget such as $k=6$ without introducing significant quantization error. By constraining the number of terms to $k=6$ across the $g=3$ values, TR is are able to ensure a tighter processing bound compared to 4-bit QT. Specifically, assuming each data value has up to $7$ term, the maximum number of term pairs with TR is reduced to $7 \times k = 42$, which is smaller than 4-bit QT of $7 \times 4 \times 3 = 84$ by a factor of $2\times$.

\subsection{Term Pair Reduction for Term Revealing Groups}
\label{sec:computing-term-revealing}

To more formally quantify the term pair reduction due to TR, suppose for a group size of $g = 3$ that the group budget is $k$ and the receding water algorithm reveals $k_1$, $k_2$ and $k_3$ number of terms for weight values $w_1$, $w_2$ and $w_3$, respectively, with $k = k_1 + k_2 + k_3 $. Suppose further that $x_1, x_2$ and $x_3$ have $r_1, r_2$ and $r_3$ terms, respectively. (For example, $k_1 = 2$, $k_2 = 3$, $k_3 = 1$, and $r_1 = 2$, $r_2 = 4$, $r_3 = 3$.) Then, with TR, the total number of term pairs to be processed for the dot product computation between $x$ and $w$ is

\centerline{
$r_1k_1 + r_2k_2 + r_3k_3 \leq 7 \times (k_1 + k_2 + k_3) = 7 \times k$
}

In reality, since most weights and data require significantly fewer than the maximum allotted number of power-of-two terms, for most groups, dot products will complete the computation below this bound, as discussed earlier in relation to Figure~\ref{fig:term-group-dist}. In this sense, TR can be viewed as shifting this upper bound from $7\times7\times g$ terms per group in the baseline case (7 terms for both weights and data) to $7\times k$ terms per group, where $k \ll 7\times g$. In Section~\ref{sec:term revealing-hw}, we utilize this significantly reduced upper bound enabled via TR to implement tightly synchronized processor arrays for DNN inference.

\subsection{Relationship Between Group Size and Group Budget}
\label{sec:config-group-budget}

TR budgets $k$ terms for a group of size $g$. Let $k = \alpha \times g$ for some $\alpha$, where $\alpha$ is the average number of terms budgeted for each value in the group. Recall from Figure~\ref{fig:values-vs-terms-dist} that $79\%$ of weight values are represented in 3 or fewer terms. This means that as the group size $g$ increases, the average number of budgeted terms per value approaches the mean of the weight term distribution. For the weight term distribution in Figure~\ref{fig:values-vs-terms-dist}, the mean is only $2.46$ terms per values, even though some values have as many as $7$ terms. Practically, this means that a larger group size allows for a smaller relative term budget $k$ which is close to the mean, as it becomes increasingly unlikely that many groups have more than $k$ terms.

\subsection{Bounding Truncation-induced Error in Dot Products}
\label{sec:bound}

TR strives to minimize truncation-induced relative error $\sigma$. Suppose that $2^i$ is the water line determined by TR under a given group budget. That is, terms smaller than $2^i$ are truncated. Then, for a group size $g$ and $\alpha = 1.5$, kept terms have value at least $g \times 2^i + \frac{g}{2} \times 2^{i+1}$, or $g \times 2^{i+1}$, and truncated terms have value at most $g \times (2^{i-1}+2^{i-2}+ \dots +2^0)$, or $g \times (2^i-1)$. We have $\sigma = \frac{\textrm{truncated\_terms}}{\textrm{kept\_terms} + \textrm{truncated\_terms}} \le \frac{\textrm{truncated\_terms}}{\textrm{kept\_terms}} \le \frac{2^i-1}{2^{i+1}} \le \frac{1}{2}$. Larger $\alpha$ results in a reduced upper bound on $\sigma$. 

We provide a bound on the relative error introduced by TR in truncated dot products between weights and data. For a given group of data ($x_1$, $x_2$, $x_3$), the dot product over the group is $w_1 x_1 + w_2  x_2 + w_3  x_3$ where $w_1$, $w_2$ and $w_3$ are corresponding weights of the filter. Each $w_i$ values may be positive, negative or zero, while $x_i$ data values are non-negative. For simplicity, we assume here that all $w_i$ are positive while noting the result also holds when they are all negative. After TR, each $x_i$ is replaced with a truncated $x_i^{\prime}$ in the dot product computation. Let $\sigma_i$ denote the relative error of $x_i^{\prime}$ induced by TR, i.e., $x_i^{\prime} = x_i  (1-\sigma_i) = x_i - x_i  \sigma_i$. Then, the dot product result $y$ with $x_i^{\prime}$ can be decomposed as follows:

\vspace{-1.2em}
{\setlength{\mathindent}{0cm}
\begin{equation*}
\begin{split}
y &= w_1  x_1^{\prime} + w_2  x_2^{\prime} + w_3 x_3^{\prime} \\
       &= w_1(x_1 - x_1 \sigma_1) + w_2(x_2 - x_2  \sigma_2) + w_3(x_3 - x_3 \sigma_3) \\
       &= w_1  x_1 + w_2  x_2 + w_3  x_3 - (w_1  x_1  \sigma_1 + w_2  x_2  \sigma_2 + w_3  x_3  \sigma_3)
\end{split}
\end{equation*}}
\vspace{-0.4em}

\noindent Therefore, the relative error of the dot product with truncated values as an approximation to the original dot product is:

\vspace{-0.8em}
\begin{equation*}
\hspace*{1.5cm} \frac{w_1  x_1  \sigma_1 + w_2  x_2  \sigma_2 + w_3  x_3  \sigma_3}{w_1  x_1 + w_2  x_2 + w_3  x_3}
\end{equation*}
\vspace{-0.8em}

\noindent Suppose that, as described above, by TR we can assure that
$\sigma_i \le \sigma$ for $i = 1, 2, 3$. Then,

\vspace{-0.8em}
{\setlength{\mathindent}{0cm}
\begin{equation*}
\begin{split}
&\frac{w_1  x_1  \sigma_1 + w_2  x_2  \sigma_2 + w_3  x_3  \sigma_3}{w_1  x_1 + w_2  x_2 + w_3  x_3}
\le \frac{w_1  x_1  \sigma + w_2  x_2  \sigma + w_3  x_3  \sigma}{w_1  x_1 + w_2  x_2 + w_3  x_3} \\
\end{split}
\end{equation*}}
\vspace{-0.8em}

\noindent Thus, the relative error in the computed dot products $w_1  x_1^{\prime} + w_2  x_2^{\prime} + w_3  x_3^{\prime}$ is bounded by $\sigma$.

\section{Hybrid Encoding for Shortened Expressions}
\label{sec:power-of-two}

In this section, we present Hybrid Encoding for Shortened Expressions (HESE), a signed power-of-two encoding which produces a signed digit representation (SDR) with the minimum number of terms for binary input. HESE complements TR by reducing the number of terms per value in a group before TR is applied across the group.

\subsection{Signed Digit Representations}
\label{sec:signed-rep}

Signed Digit Representations (SDRs) are a type of positional encoding system, where each position can have a coefficient of $\{-1, 0, 1\}$ as opposed to only $\{0, 1\}$ in a conventional binary encoding. Avizienis proposed the use of SDRs in bit-parallel circuits to remove carry-propagation chains in additions and multiplications~\cite{avizienis1961signed}. Drake et al. proposed a similar approach involving carry-free addition and subtraction in an optical computing regime~\cite{drake1986photonic}. More commonly, booth radix-4 encoding~\cite{booth1951signed} has been used to convert binary representations with only positive power-of-two terms (\eg~$30 = 2^4 + 2^3 + 2^2 + 2^1$) into shorter representations with both positive and negative power-of-two terms (\eg~$30 = 2^5 - 2^1$). 

Booth radix-4 encoding bounds the number of power-of-two terms in an n-bit value to $\frac{n}{2} + 1$~\cite{booth1951signed}. This is utilized in the design of efficient booth multipliers to provide a smaller bound on the amount of computation required for any pair of n-bit values. This bound is necessary for synchronization purposes across multiple processing elements (\eg~systolic arrays). However, booth encoding does not lead to the minimum-length SDR in the case of a number with an isolated zero. For instance, $27$ ($11011$ in binary) will be converted into $10\bar{1}10\bar{1}$ in Booth, while the minimum-length encoding is $100\bar{1}0\bar{1}$.\footnote{Here, $\bar{1}$ represents a term with a negative coefficient, such as $-2^2$.}

There has been other prior work on developing algorithms to convert binary to minimum-length SDR. Jedwab et al. proposed such an algorithm, with a subsequent proof that the generated SDR has the minimum-possible number of terms~\cite{jedwab1989minimum}. However, the proposed algorithm is not amenable to efficient hardware implementations as written, since it requires multiple passes through the input. In the next section, we present an efficient one-pass two-bit encoding method called HESE for producing SDRs that still achieves the minimum number of terms.

\subsection{Overview of HESE}
\label{sec:hese}

\begin{figure}
\centering
\begin{minipage}{0.4\columnwidth}
  \vspace{0pt}
  \centering
  \includegraphics[width=\columnwidth]{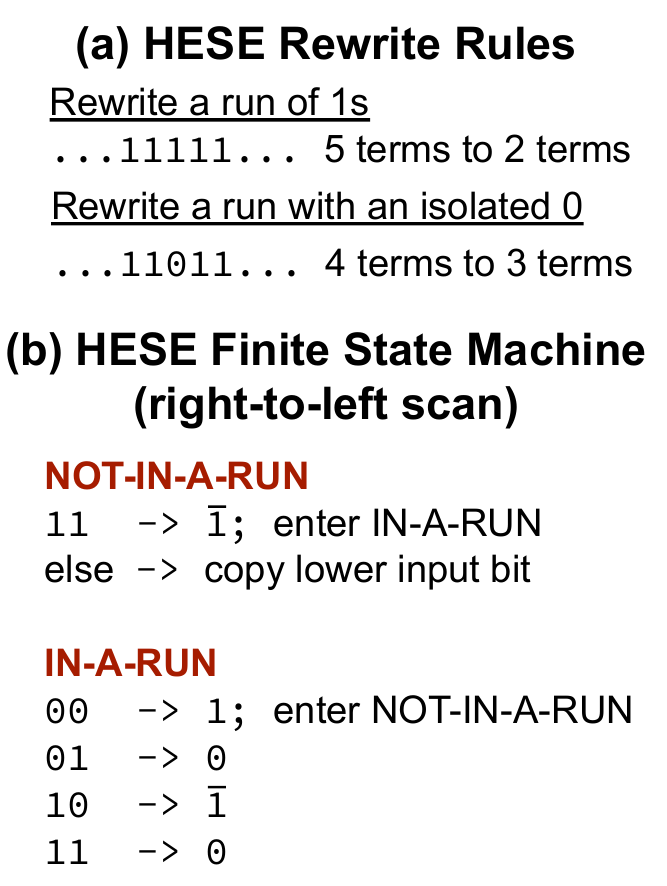}
\end{minipage}%
\begin{minipage}{0.55\columnwidth}
  \vspace{0pt}
  \centering
  \includegraphics[width=\columnwidth]{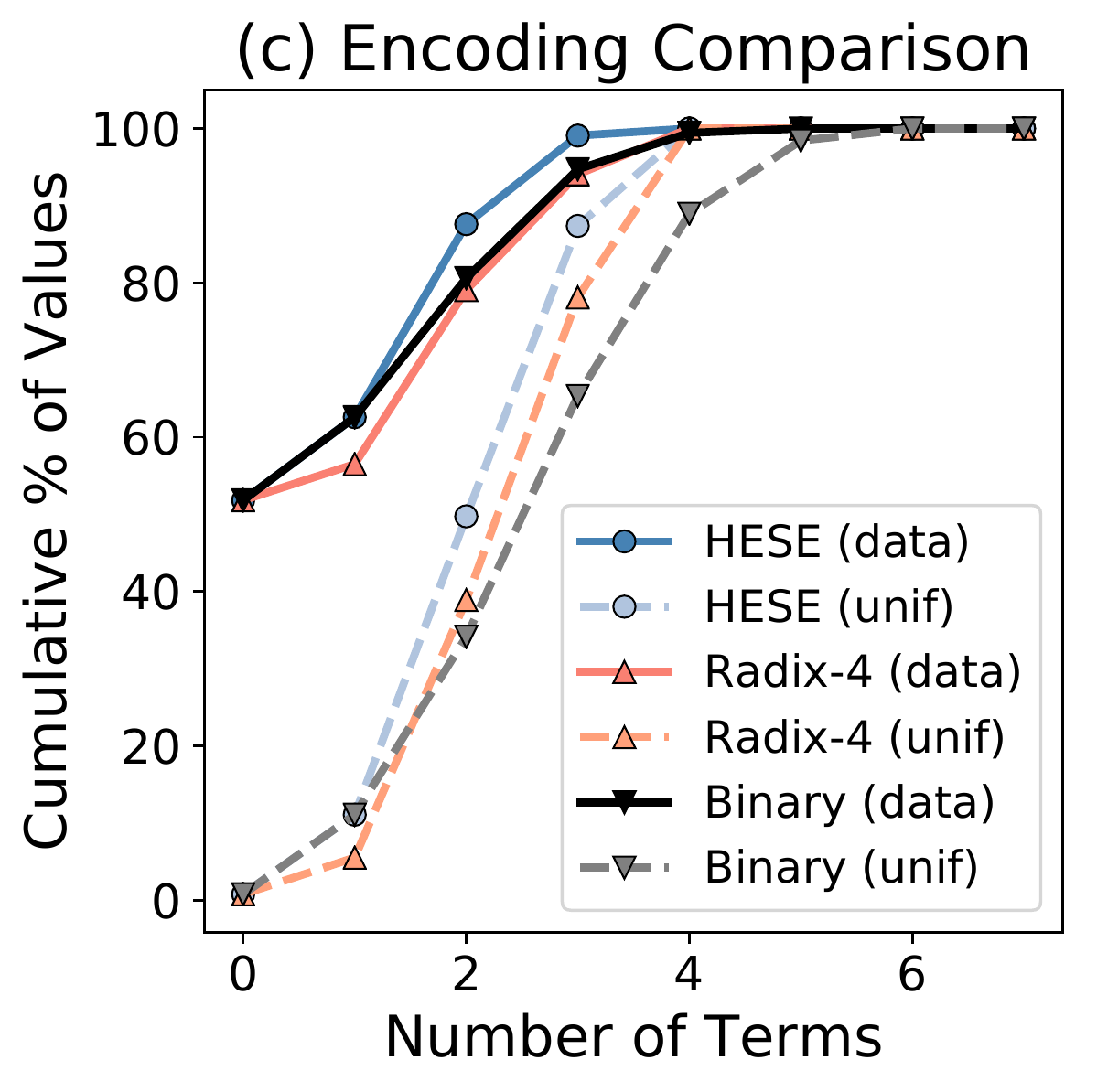}
\end{minipage}%
{\phantomsubcaption\label{fig:hese-encoding}}
{\phantomsubcaption\label{fig:hese-fsm}}
{\phantomsubcaption\label{fig:term-cdf}}
\caption{(a) HESE rewrite rules for converting binary encodings into SDRs. (b) The HESE finite state machine, which operates on two input bits and outputs one digit per state transition. (c) The SDRs generated by HESE require fewer terms than both binary and radix-4 encodings for 8-bit values over DNN data (data) and a uniform distribution (unif).}
\label{fig:hese-all}
\end{figure}

HESE is a hybrid encoding method that combines Booth, which efficiently handles strings of $1$s, with an additional rules for an isolated $0$ surrounded by at least two $1$s on each side. The first rewrite rule in Figure~\ref{fig:hese-encoding} shows that this encoding reduces a sequence of 5 $1$s, such as ($11111$) to only two terms ($10000\bar{1}$). The second rule shows how a sequence with an isolated $0$, such as $11011$ is rewritten as $100\bar{1}0\bar{1}$, which has only 3 terms. Isolated $1$s in the input remain $1$s in the output. Figure~\ref{fig:hese-fsm} provides a finite state machine for HESE. It begins in the NOT-IN-A-RUN mode and each state corresponds to two bits in the input sequence. On state transitions, it consumes a single bit of the input and outputs a single signed digit. Transitions from the NOT-IN-A-RUN mode to the IN-A-RUN mode are triggered by observing a $11$ in the input (denoting a run of at least two $1$s). Likewise, transitions from IN-A-RUN mode back to NOT-IN-A-RUN mode occur when a $00$ sequence is observed (or the input is consumed).

HESE can also be extended to convert non minimum-length SDRs into minimum-length SDRs. Basically, by replacing adjacent mixed-sign nonzero terms, $+-$ or $-+$, with a nonzero term and a zero term, we end up with strings of $1$s or strings of $-1$s. Like before, the two rules in Figure~\ref{fig:hese-encoding} can rewrite these strings as well as strings with isolated $0$s to derive the minimum-length SDR. However, in this work, we only use HESE to convert binary to minimum-length SDRs.

\subsection{Reducing Number of Terms per Encoding}
\label{sec:truncated-power-of-two}

HESE encodings have strictly equal or fewer terms than binary and Booth radix-4. Figure~\ref{fig:term-cdf} shows the number of terms required for these encodings across two distributions of values: data values from ResNet-18 and values drawn from a uniform distribution over the same range as the data. The x-axis is the number of terms required to represent a value and the y-axis is the cumulative percentage of values that are represented within a given number of terms. HESE outperforms both Booth and binary across both distributions of values.

As expected, Booth leads to more compact representations than binary for values drawn from the uniform distribution. However, most of the reduction in terms comes from larger values in the 8-bit range (with many 1s), which occur much less frequently for the data, as depicted in Figure~\ref{fig:values-vs-terms-dist} (bottom). Therefore, radix-4 performs equal or worse than binary for the distribution of data values we are interested in. By comparison, when applying HESE on data, $99\%$ of values are represented in 3 or fewer terms. Practically, this means we can use 3 power-of-two terms for both weights and data.

\section{Hardware Design for Efficient Term Revealing}
\label{sec:term revealing-hw}

\begin{figure}
    \centering
    \includegraphics[width=\columnwidth]{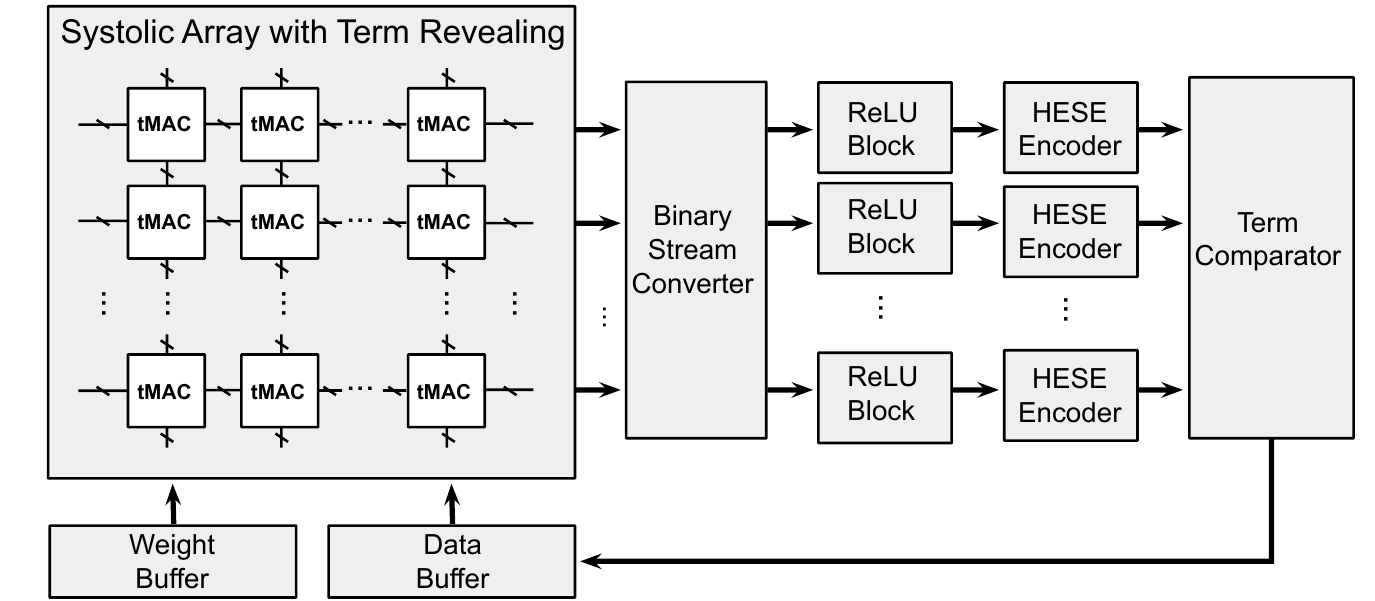}
    \caption{The term revealing system design.}
    \label{fig:system}
\end{figure}

In this section, we present our hardware design for TR-based quantization. Figure~\ref{fig:system} provides an overview of the TR system design, consisting of the following components: (1) weight and data buffers which store DNN layer weights and input/intermediate data, (2) a systolic array which performs dot products between weights and data using term MACs (tMAC) described in Section~\ref{sec:tMAC-design}, (3) a binary stream converter to convert systolic array output into binary representation (Section~\ref{sec:relu-block}), (4) a ReLU block (Section~\ref{sec:relu-block}), (5) a HESE encoder (Section~\ref{sec:hese-encoder}) to convert the binary representations to shorter SDRs, and (6) a term comparator (Section~\ref{sec:comparator}) which applies TR by selecting the top $k$ terms in a group. In Section~\ref{sec:tr-systolic}, we first give some high-level reasoning on how tMAC can save computation.

\subsection{High-level Comparison Between Bit-parallel MAC (pMAC) and Term MAC (tMAC)}
\label{sec:tr-systolic}
To help understand the inherent advantage of TR, we provide a high-level argument on how our proposed term MAC (tMAC) saves a significant amount of work over a conventional parallel MAC (pMAC). Here, we define \textit{work} as the amount of computation, including both arithmetic and bookkeeping operations, which are performed per group. The work incurred by a method largely determines the energy, area, and latency of its implementation.

\begin{figure}
    \centering
    \includegraphics[width=\columnwidth]{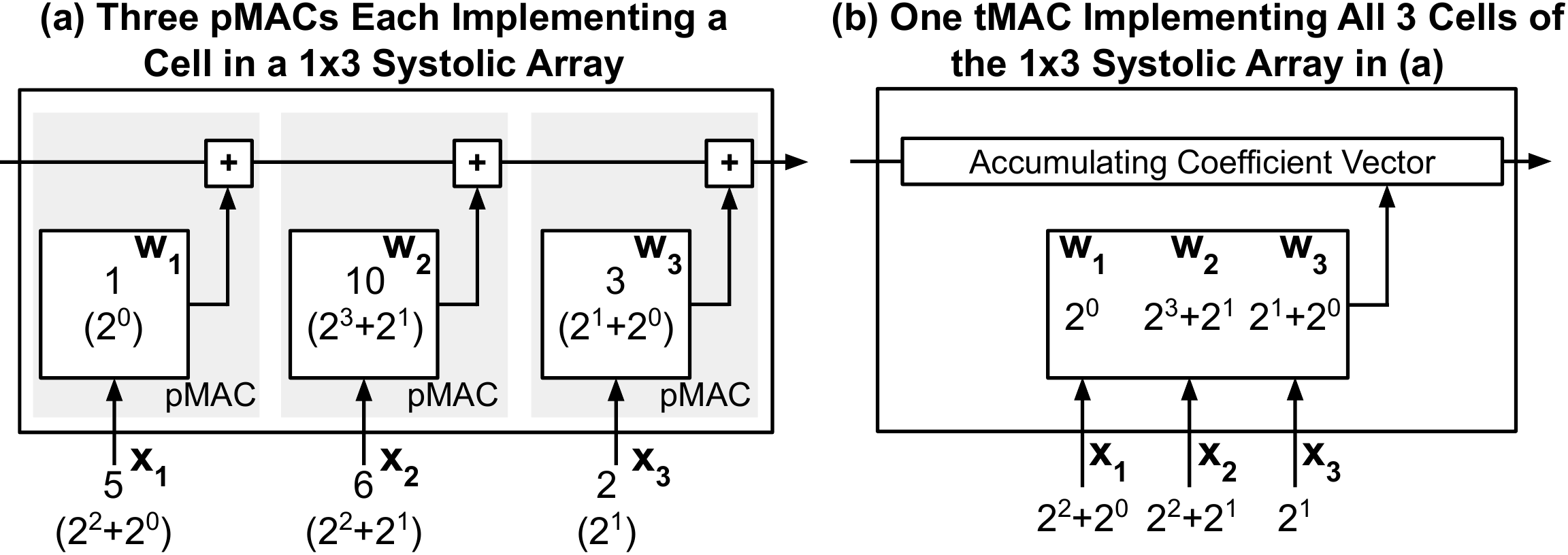}
    {\phantomsubcaption\label{fig:pmac-tmac-a}}
    {\phantomsubcaption\label{fig:pmac-tmac-b}}
    \caption{(a) A $1\times3$ systolic array where each of the three systolic cells is a conventional bit-parallel MAC (pMAC) which performs an 8-bit multiplication between weights ($w$) and data ($x$) values and a 32-bit accumulation each systolic array cycle. (b) The proposed term MAC (tMAC) processes all term-pair multiplications (e.g., $2^2 \cdot 2^1$), for the same systolic array cycle, across a group of weight and data values (group size
    $g=3$ here) in a bit-serial fashion. For a group budget of $k$ and $s$-term data, the number of term-pair multiplications is bounded by $k\cdot s$. 
    Here, with $k = 6$ and $s = 2$, it is 8 ($< 6 \cdot 2 = 12$).}
    \label{fig:pmac-tmac}
\end{figure}

To be concrete, we study a $1\times3$ 1D systolic array of 3 cells, as depicted in Figure~\ref{fig:pmac-tmac-a},
for the processing of groups of 3 data values ($x_1$, $x_2$, $x_3$) in computing their dot products with weights ($w_1$, $w_2$, $w_3$) pre-stored in the systolic array. To provide a baseline for comparison, we consider a conventional implementation, where each cell is a pMAC performing an 8-bit bit-parallel multiplication, $w \times x$, and a 32-bit accumulation adding an intermediate $y$ to the computed $w \times x$. In comparison, Figure~\ref{fig:pmac-tmac-b} depicts a tMAC-based implementation which significantly reduces the work by only processing available term pairs.  The number of terms is relatively  small due to high bit-level sparsity generally presented in CNN weights and data. This comparison result applies to a general 2D systolic array, which is a stack of 1D systolic arrays. \textbf{In Section~\ref{sec:tmac-pmac-comparison}, we show how this analysis translates to realized performance on an FPGA with a group size of $g = 8$.}

For this illustrative analysis, we assume that tMAC uses a TR group of size $g = 3$ and budget $k = 6$ for weight values, and $s = 2$ leading terms for data values under HESE (Section~\ref{sec:hese}). Thus, for weights, each value in a group uses on average $\alpha = 2$ terms. As we show in Table~\ref{table:energy-eff-comparison}, under similar settings, TR will incur a minimum decrease in classification accuracy (e.g., less than $0.15\%$) when dropping lower-order terms exceeding the group budget $k$ across multiple CNNs.

Our analysis on work proceeds as follows. A conventional pMAC implementation of a single systolic cell incurs 7 8-bit additions for the multiplication $w \times x$ and 1 32-bit accumulation operation for $y + w \times x$. Therefore, the pMAC implementation of a $1\times3$ 1D systolic array with three cells requires 21 8-bit additions and 3 32-bit accumulation operations. In contrast, a tMAC implementation incurs significantly less work. Specifically, it uses at most 12 3-bit additions on exponents of power-of-two terms (weight and data exponents are both less than 8) for term-pair multiplications. (Recall that we assume $k=6$ and $s = 2$ terms for data values, as depicted in Figure~\ref{fig:pmac-tmac-b}). The updating of accumulating coefficient vector (discussed in detail in the next section) requires bookkeeping operations for bit alignment, etc., with work we assume is no larger than the equivalent 12 3-bit additions. Thus, tMAC substantially reduces work compared to pMAC, that is, 24 3-bit additions vs. 21 8-bit additions plus 3 32-bit accumulations.

\subsection{Term MAC (tMAC) Design}
\label{sec:tMAC-design}

\begin{figure}
    \centering
    \includegraphics[width=0.9\columnwidth]{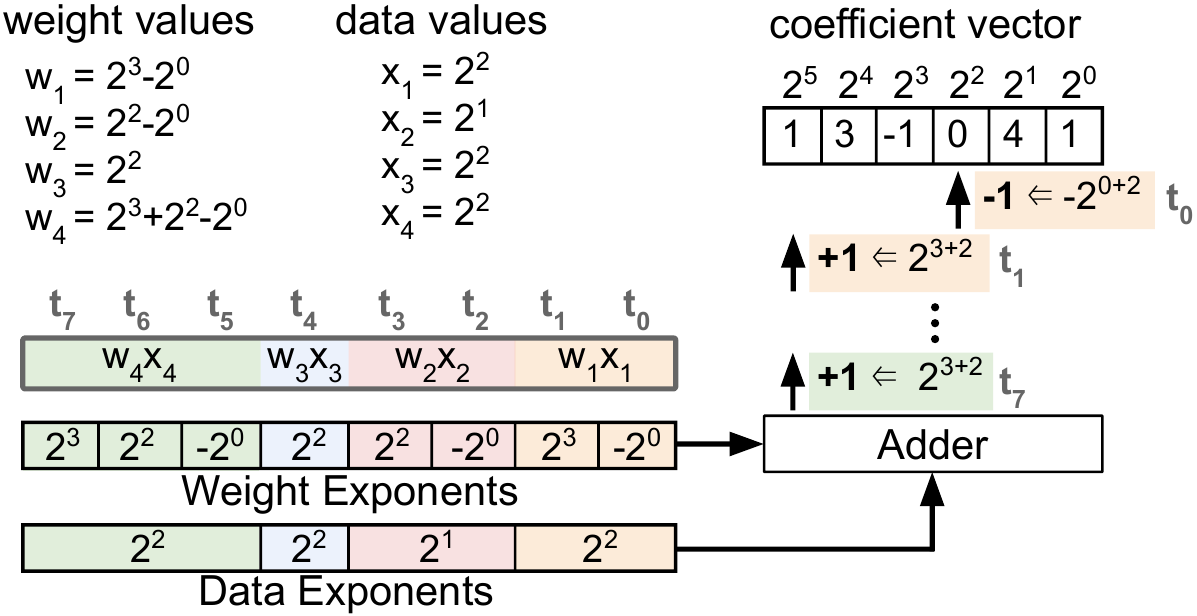}
    \caption{Term pair multiplication for a dot product across a group of 4 weight and data values over 8 cycles ($t_0$ to $t_7$).}
    \label{fig:exponent-adder}
\end{figure}

The term MAC (tMAC) performs dot products between a data and weight vector of group size $g$ by multiplying all term pairs. Figure~\ref{fig:exponent-adder} illustrates how these term pair multiplications in tMAC are performed for a group of size $g=4$ and a group budget $k=8$. In this example, TR ensures that there are $8$ or fewer terms across all weight values in the group. For illustration simplicity, assume all data values can be represented with a single term (in our implementation there are as many as $3$ terms per data value). Under these assumptions, $8$ term pair multiplications are performed and the results are added to a coefficient vector depicted in the upper right of Figure~\ref{fig:exponent-adder}.

The coefficient vector stores the current partial result of the dot product as coefficients for each power of two. In this example, the coefficients are set to ($1$, $3$, $-1$, $0$, $4$, $1$), which represents a value of $1\times2^5 + 3\times2^4 - 1\times2^3 + 0\times2^2 + 4\times2^1 + 1\times2^0 = 81$. For the first term pair in the figure, $(-2^0, +2^2)$ in $w_1x_1$, the coefficient for $2^2$ is decremented by 1 as the signs of the terms differ. Once all exponent additions are completed for a dot product, the coefficient vector is reduced to a single value. As the largest term is $2^{7}$, assuming 8-bit uniform quantization, the largest term pair is $2^{7}\times 2^{7} = 2^{14}$. Therefore, the coefficient vector has a length of $15$ in order to store all possible term pair results from $2^{0}$ to $2^{14}$. To ensure overflow is not possible for dot products of length as large as $4,096$, each element in the coefficient vector is 12 bits. 

\begin{figure}
    \centering
    \includegraphics[width=\columnwidth]{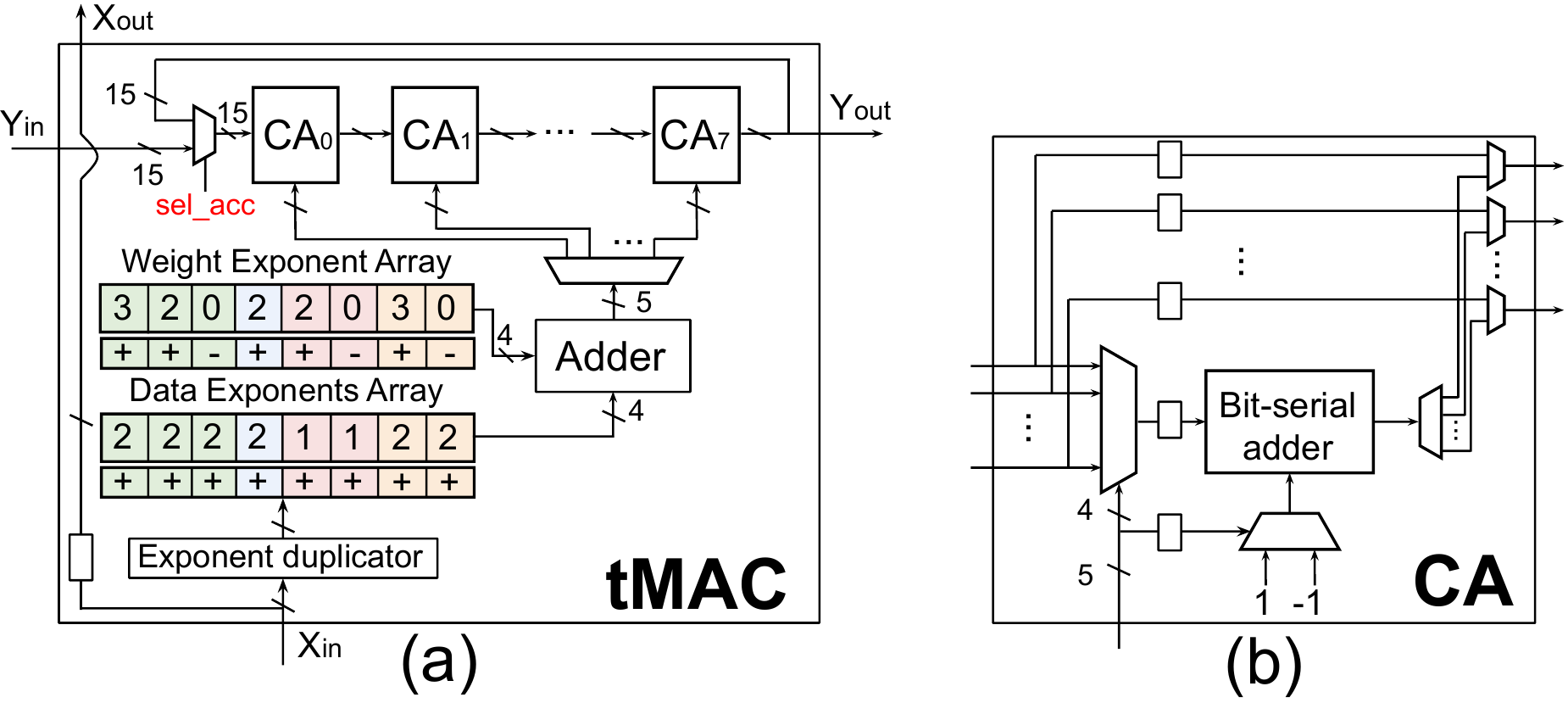}
    {\phantomsubcaption\label{fig:hardware-components-a}}
    {\phantomsubcaption\label{fig:hardware-components-b}}
    \caption{(a) The term MAC (tMAC) performs term pair multiplications between data and weight terms for a group of values. (b) A coefficient accumulator (CA) takes the adder result and add or subtract 1 from the corresponding coefficient.}
    \label{fig:hardware-components}
\end{figure}

The hardware design of tMAC is shown in Figure~\ref{fig:hardware-components-a}. The exponents for term pairs are stored in data and weight exponent arrays, with the sign of each term stored in the parallel arrays with one bit per term. For instance, the term $-2^2$ would store a $2$ in the exponent array and a minus ($-$) in the sign array. The yellow, red, blue, and green colors denote the term pair boundaries for each data $\times$ weight multiplication. The exponent duplicator takes in data exponents and duplicates them based on the number of weight exponents in each value. Each cycle, a pair of exponents from these two arrays are passed into the adder, which computes the sum of the exponents, sets the sign, and sends to the result to a \textit{coefficient accumulator} (CA) (Figure~\ref{fig:hardware-components-b}) within one cycle. Therefore, to process a group with 8 term pairs takes 8 cycles in total. 

The CAs perform bit-serial addition between the coefficient vector and the output of the exponent adder in the tMAC. Due to the bit-serial design, the number of CAs must match the size of the data and weight register arrays (8 in this example) in order to maintain synchronization across the cells of the systolic array. At each cycle, one of the eight CAs takes the sum of two exponents from the adder, and adds/subtracts 1 to/from the corresponding coefficient. In our implementation, each tMAC can choose to reuse the current coefficient vector or take the new coefficient vector from its neighboring cell via the selection signal $sec\_acc$, as depicted in Figure~\ref{fig:hardware-components-a}. 

\subsection{Binary Stream Converter and ReLU Block}
\label{sec:relu-block}
The binary stream converter takes coefficient vectors, output from the systolic array, and transforms them into a binary format by multiplying each element of the coefficient vector with the corresponding power-of-two term then summing over the partial results. The outputs of the binary stream converter are sent to the ReLU block in a bit-serial fashion. Using a two's complement representation for the outputs, the sign can be determined by detecting the most significant bit of the output streams. The ReLU block buffers all the lower bits until the MSB arrives. Then, it outputs zero if the sign of the MSB indicates that the value is negative; otherwise it outputs the original bit stream.  

\subsection{HESE Encoder}
\label{sec:hese-encoder}
The HESE encoder produces two bit streams, which represent the magnitude and sign of each power-of-two term, respectively. For example, for a bit-serial input of $31 = 00011111$, the HESE encoder will produce two output streams: $00100001$ (magnitude) and $00000001$ (sign), to indicate $31 = 2^{5}-2^{0}$. The HESE encoder is implemented with the finite state machine in Figure~\ref{fig:hese-fsm}.

\subsection{Term Comparator}
\label{sec:comparator}
\begin{figure}
    \centering
    \includegraphics[width=\columnwidth]{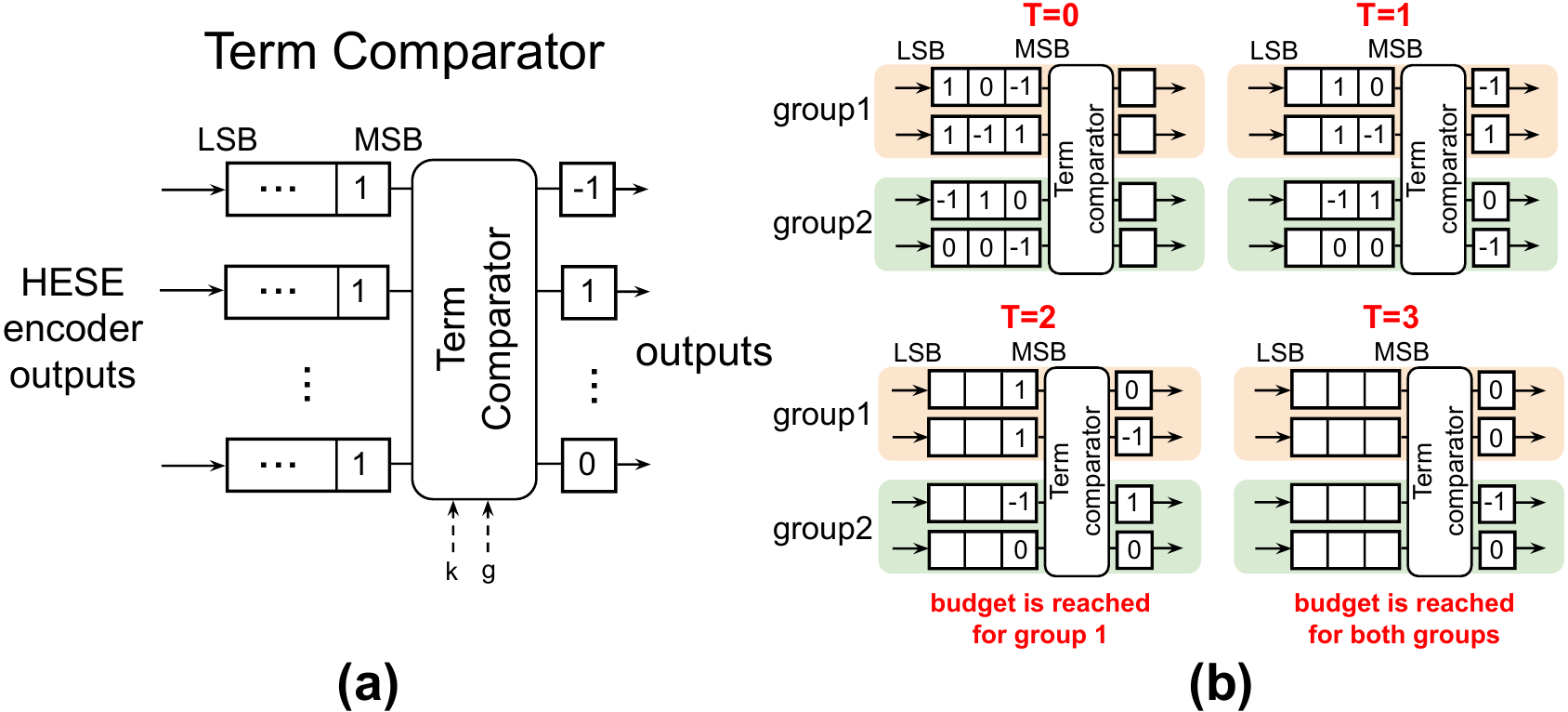}
    {\phantomsubcaption\label{fig:truncation-array-view-a}}
    {\phantomsubcaption\label{fig:truncation-array-view-b}}
    \caption{(a) The design of the term comparator which implements term revealing. The term comparator takes the group size $g$ and group budget $k$ as inputs, counts the total number of terms within each group, and set the corresponding terms to zeros once the group budget is reached. (b) An example of term comparator operating on two groups. At T=2, the group budget is reached for group 1 and all the remaining terms are pruned. At T=3, group budget is also reached for group 2.}
    \label{fig:truncation-array-view}
\end{figure}

The term comparator in Figure~\ref{fig:truncation-array-view-a} selects the top $k$ terms from the outputs of every $g$ consecutive HESE encoders, where $k$ and $g$ are the group budget and group size, respectively. Figure~\ref{fig:truncation-array-view-b} shows the operation of term comparator on the outputs of four HESE encoders, the HESE encoder outputs are divided into two groups, where each group has a group size $g=2$ and group budget of $k=3$. The inputs enter the term comparator in a reverse order such that their most significant bits (MSB) enter the term comparator first. Each cycle, the term comparator counts the total number of nonzero bits encountered so far, and truncates the remaining low-order terms once the group budget $k$ is reached for a group.

\begin{figure}
    \centering
    \includegraphics[width=\columnwidth]{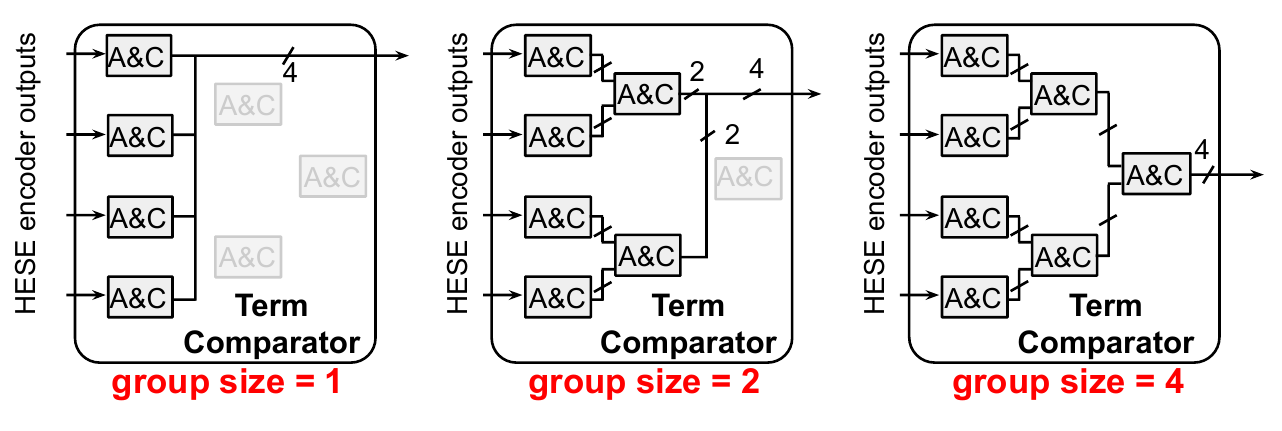}
    \caption{Configurations of term comparator under different group sizes.}
    \label{fig:reconfiguration-comparator}
\end{figure}

The term comparator contains multiple accumulate and compare (A\&C) blocks which are arranged into a tree structure. Each A\&C block takes a single input bit stream and counts the total number of nonzero bits in this stream. Figure~\ref{fig:reconfiguration-comparator} show how the A\&C blocks can be reconfigured for different group sizes $g$. For $g=1$, the A\&C blocks on the first level of the tree will compare the number of nonzero bits in their input stream against the group budget $k$, and truncate each stream accordingly. If the group size is larger than 1 (e.g., 2), each A\&C block in the first level of the tree will forward its input stream together with the nonzero bit count to its parent A\&C block. The parent A\&C block then operates on these two streams in a similar fashion to its children. The tree architecture allows for minimal changes to the term comparator under different group sizes, which leads to a low reconfiguration overhead and maximum level of hardware reuse.

\subsection{Memory Subsystem}
\label{sec:memory-encoding}

Our memory subsystem consists of a data and weight buffer. The data buffer holds the term exponents and signs for both the input and result data of the current layer, and the weight buffer holds the term exponents and signs of the weights for each group. For the weight buffer, we use double buffering to prefetch the next weight tile from the off-chip DRAM so that the computation of the systolic array can overlap with the traffic transfer from the off-chip DRAM to weight buffer. Note that TR does not reduce the storage complexity of the model, as each weight is stored in an 8-bit fixed-point format.

\subsection{FPGA Reconfiguration for QT and TR}
\label{sec:hardware-reconfiguration}
Our TR system system can be easily reconfigured for different group sizes $g$ and group budgets $k$, in order to adapt to dynamic requirements on group size and group budget during inference with a negligible delay. In addition, our system can also supports conventional quantization (QT) by performing power-of-two operations with binary representations. Since QT does not require TR or HESE encoding, the term comparator and HESE encoder can be turned off by using clock gating to reduce power consumption. Table~\ref{tab:hardware-components} summarizes all of the control registers which need to be modified when switching between TR and QT. The switching process only takes several clock cycles (i.e., within 100ns for our FPGA implementation). 

\begin{table}
\centering
\caption{Control registers for supporting QT and TR.}
\begin{adjustbox}{width=\columnwidth,center}
\begin{tabular}{|c|c|c|}
 \hline
 &  \begin{tabular}{@{}c@{}}Uniform quantization (QT) \end{tabular} &  \begin{tabular}{@{}c@{}} Term revealing (TR)\\ \end{tabular}    \\ \hline
\begin{tabular}{@{}c@{}}HESE\_ENCODER\_ON\\(1 bit)\end{tabular}   &   \begin{tabular}{@{}c@{}}HESE encoder is turned\\off by setting this bit to 0\end{tabular} &\begin{tabular}{@{}c@{}}HESE encoder is turned\\on by setting this bit to 1 \end{tabular}     \\ \hline
\begin{tabular}{@{}c@{}}COMPARATOR\_ON\\(1 bit)\end{tabular} & \begin{tabular}{@{}c@{}}term comparator is turned\\off by setting this bit to 0\end{tabular} & \begin{tabular}{@{}c@{}}term comparator is turned\\on by setting this bit to 1\end{tabular} \\\hline

\begin{tabular}{@{}c@{}}QUANT\_BITWIDTH\\(4 bit)\end{tabular} & \begin{tabular}{@{}c@{}}quantization bitwidth\\ used for QT\end{tabular} & \begin{tabular}{@{}c@{}}quantization bitwidth\\ used for TR\end{tabular}  \\\hline
\begin{tabular}{@{}c@{}}DATA\_TERMS\\(4 bit)\end{tabular}  & \begin{tabular}{@{}c@{}} same as the quantization\\ 
bitwidth for QT \end{tabular} & \begin{tabular}{@{}c@{}}Maximum number of power-\\of-two terms in data for TR\end{tabular}   \\\hline

\begin{tabular}{@{}c@{}}GROUP\_SIZE\\(3 bit)\end{tabular} &\begin{tabular}{@{}c@{}}group size is set\\to 1 for QT\end{tabular} & \begin{tabular}{@{}c@{}}group size is between\\2 to 8 for TR\end{tabular}\\\hline

\begin{tabular}{@{}c@{}}GROUP\_BUDGET\\(5 bit)\end{tabular} &  \begin{tabular}{@{}c@{}} group budget is the same as \\quantization bitwidth for QT \end{tabular}  & \begin{tabular}{@{}c@{}}group budget can be up\\to $8\times 3= 24$ for TR \end{tabular}  \\ \hline
\end{tabular}
\end{adjustbox}
\label{tab:hardware-components}
\end{table}

\section{Term Revealing Evaluation}
\label{sec:cnn-perf}

In this section, we evaluate the performance of TR when applied to an MLP on MNIST~\cite{lecun1998mnist}, a broad range of CNNs (VGG-19~\cite{simonyan2013deep}, ResNet-18~\cite{he2016deep}, MobileNet-V2~\cite{sandler2018mobilenetv2}, and EfficientNet-b0~\cite{tan2019efficientnet}) on ImageNet~\cite{deng2009imagenet}, and an LSTM~\cite{hochreiter1997long} on Wikitext-2~\cite{merity2016pointer}. In Section~\ref{sec:tr-vs-quant-acc}, we compare TR against conventional uniform quantization (QT) on the performance (i.e., accuracy or perplexity) of these DNNs. Then, in Section~\ref{sec:group-size-accuracy}, we provide analysis on how the $\alpha$ (average number of terms) and $g$ (group size) parameters impact the classification accuracy. Next, in Section~\ref{sec:qt-tr-analysis}, we analyze the individual contribution of HESE and TR on model performance. Finally, in Section~\ref{sec:quant-error-analysis}, we show that the quantization error introduce by TR is substantially less than when using a more aggressive QT setting (e.g., 6-bit uniform quantization). 

\begin{figure*}
\centering
\begin{minipage}[t]{0.68\textwidth}
  \vspace{0pt}
  \centering
  \includegraphics[width=\columnwidth]{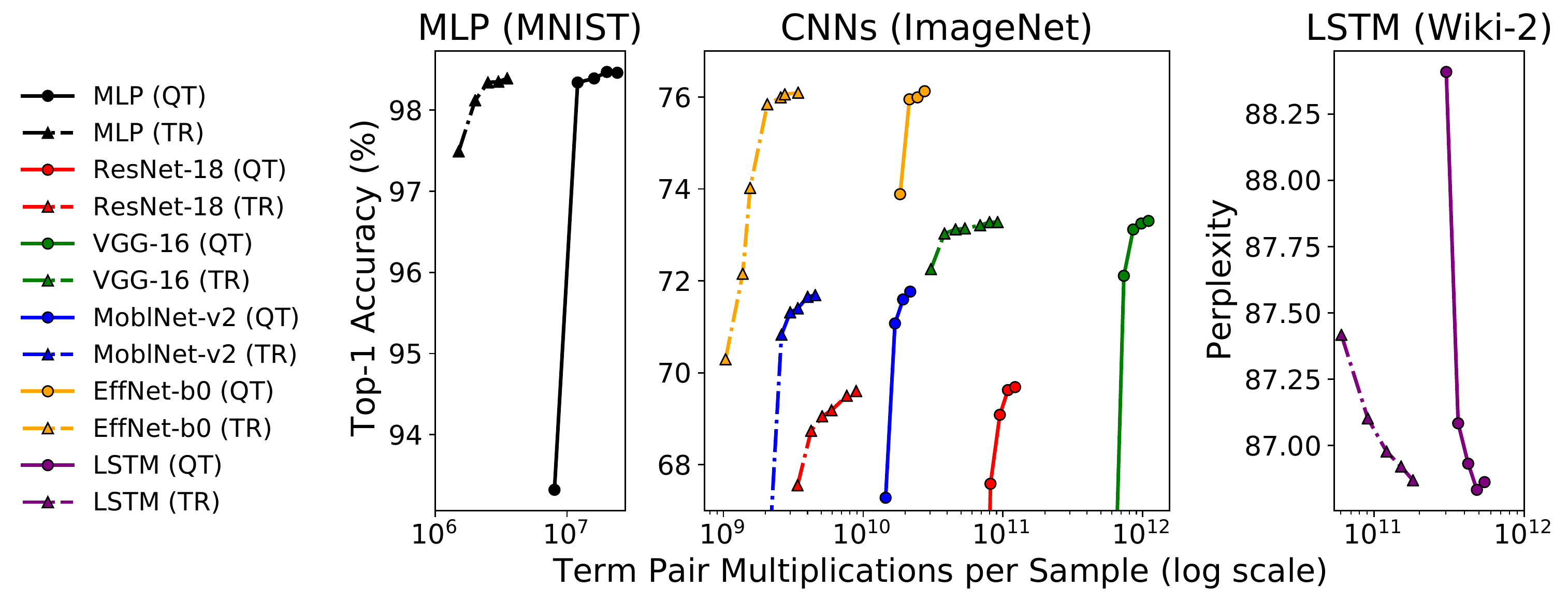}
  \caption{Comparing uniform quantization (QT) and term revealing (TR) for an MLP on MNIST (left), CNNs on ImageNet (center), and an LSTM on Wikitext-2 (right). The QT settings vary the weight bit-width (from 4 to 8 bits), while the TR settings vary $g$ (group size) and $\alpha$ (number of terms per group). TR reduces the number of term pair multiplications per sample over QT by 3-10$\times$ across the three types of DNNs.}
  \label{fig:tr-quant-acc}
\end{minipage}\hfill%
\begin{minipage}[t]{0.29\textwidth}
  \vspace{0pt}
  \centering
  \includegraphics[width=\columnwidth]{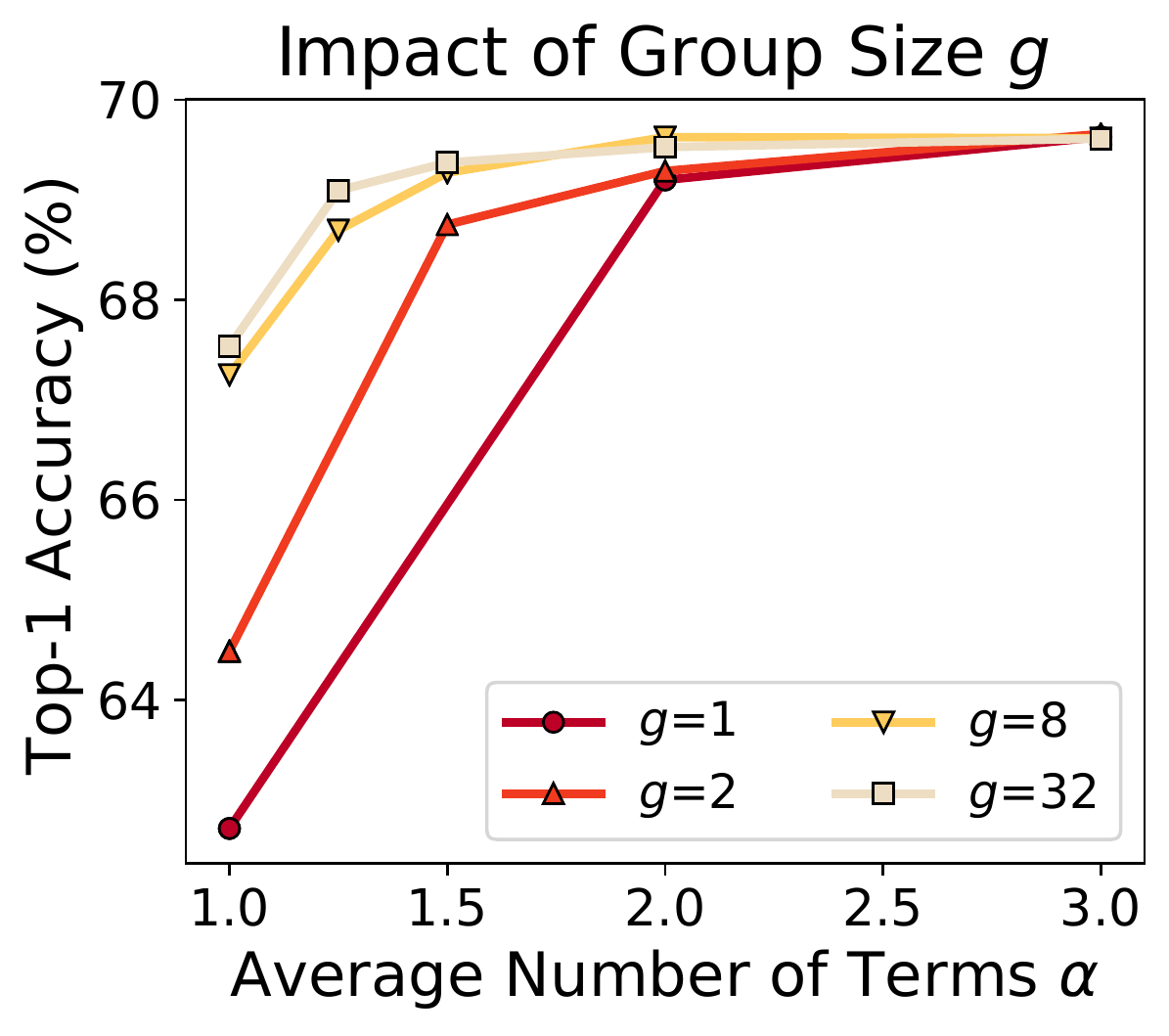}
  \caption{A larger group size $g$ improves ResNet-18 ImageNet classification accuracy for a given $\alpha$.}
  \label{fig:tr-group-size}
\end{minipage}%

\label{fig:tr-quant-comp}
\end{figure*}

To perform this analysis, we have implemented a CUDA kernel for TR which only increases the  inference runtime of a pre-trained model running on a NVIDIA 1080 Ti by under $5\%$. This means that the validation accuracy for a pre-trained CNN for ImageNet can still be obtained within several minutes. Using pre-trained models has the advantage of making parameter search (e.g., for group size $g$ and term budget $k$) simple compared to methods such as weight pruning~\cite{han2015deep} that require model retraining which takes hours or days for each setting. Before applying TR, each model is quantized from 32-bit floating-point to 8-bit fixed-point using a layerwise procedure described in~\cite{lee2018quantization}.

\subsection{Comparing Term Revealing to Uniform Quantization}
\label{sec:tr-vs-quant-acc}

Motivated by the design in Section~\ref{sec:term revealing-hw}, we are interested in minimizing the number of term pair multiplications per sample, as this directly translates to the processing latency of a sample. For the uniform quantization (QT) approach with 8-bit fixed-point weights and data, each multiplication translates to $7 \times 7 = 49$ term pair multiplications. By comparison, for term revealing (TR), the number of term pair multiplications is instead bounded by the average number of term pairs which is shared across a group of values. We show that TR gives a significant reduction (e.g., 3-10$\times$) over QT while maintaining the nearly identical performance (e.g., within 0.1\% accuracy).

\subsubsection{MLP on MNIST}
We train an MLP with one hidden layer with 512 neurons for MNIST using the parameter settings given in the PyTorch examples for MNIST\footnote{\url{https://github.com/pytorch/examples/tree/master/mnist}}. Figure~\ref{fig:tr-quant-acc} (left) shows the performance of QT and TR applied to the pre-trained MLP. TR achieves a $5\times$ reduction in number of term pair multiplications over QT while achieving a clasification accuracy of 98.4\% (compared to the 98.5\% baseline).

\subsubsection{CNNs on ImageNet}
We use pre-trained models provided by the PyTorch torchvision package\footnote{\url{https://github.com/pytorch/vision/tree/master/torchvision/models}} for VGG-16, ResNet-18, and MobileNet-v2 and a PyTorch implementation\footnote{\url{https://github.com/lukemelas/EfficientNet-PyTorch}} of EfficientNet with pre-trained models.  Figure~\ref{fig:tr-quant-acc} (center) shows the performance of TR and QT for the 4 CNNs. TR achieves a 14$\times$ reduction in term pair multiplications over QT for VGG-16, which is known to be significantly overprovisioned (e.g., amenable to quantization and pruning). Even for more recent models, which have significantly fewer parameters, such as MobileNet-v2 and EfficientNet-b0, TR is still able to achieve a 4$\times$ and 6$\times$ reduction in term pair multiplications, respectively, losing less than 0.1\% classification accuracy compared to the 8-bit QT settings. Generally, we see that more aggressive TR settings (e.g., with a reduced group budget) appears to degrade the accuracy more gracefully than more aggressive QT settings (e.g., with reduced bit-width for weight).

\subsubsection{LSTM on WikiText-2}
We train a 1-layer LSTM with 650 hidden units (i.e., neurons), a word embedding of length 650, and a dropout rate of 0.5, following the PyTorch word language model example\footnote{\url{https://github.com/pytorch/examples/blob/master/word_language_model}}. This baseline model achieves a perplexity of 86.85. Figure~\ref{fig:tr-quant-acc} (right) shows how the perplexity of the pre-trained model is impacted by QT and TR. Again, we find that TR is able to reduce the number of term pair multiplications by a significant factor of $3\times$, while achieving the same perplexity.

\subsection{Improved Term Allocation with Larger Group Size}
\label{sec:group-size-accuracy}

Figure~\ref{fig:tr-group-size} shows the classification accuracy for ResNet-18 as $\alpha$ is varied for different group sizes. As the group size increases, the variance in number of terms across values in a group shrinks, meaning that a larger group budget $k$ at a fixed $\alpha$ ratio is strictly better than a smaller $k$ for the same $\alpha$. As observed, the classification accuracy for a larger group size strictly outperforms all settings with smaller group sizes. For instance, a group size of 8 with $\alpha$ of 1 achieves a classification accuracy of 67.72\% which is 5.21\% better than a group size of 1 at the same $\alpha$ value. Note that a group size of 1 is equivalent to truncating each value to exactly $\alpha$ terms.

\subsection{Isolating the Impact of TR and HESE}
\label{sec:qt-tr-analysis}

\begin{figure}
    \centering
    \includegraphics[width=0.9\columnwidth]{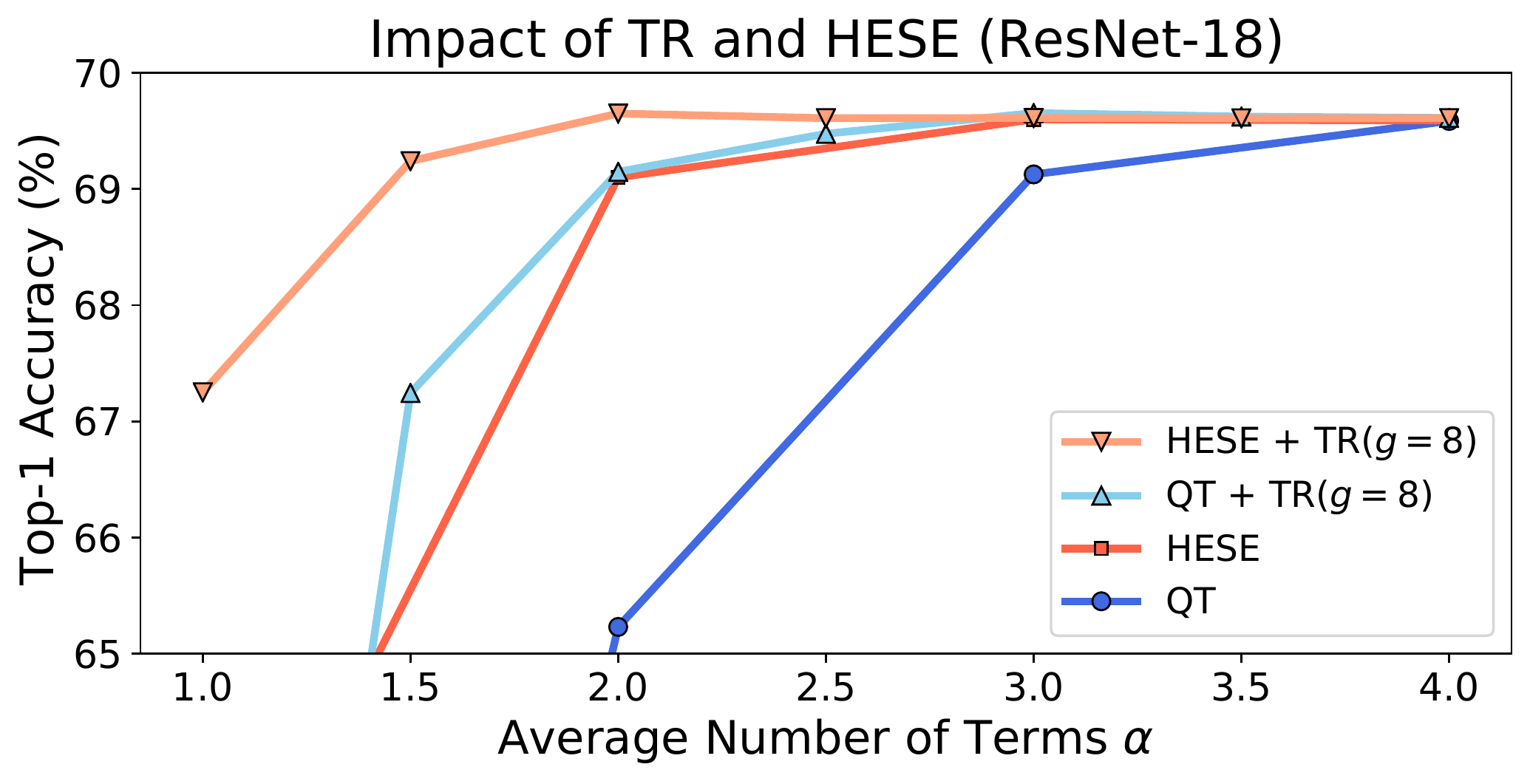}
    \caption{Measuring the individual contributions of TR and HESE in reducing number of terms while maintaining high classification accuracy.}
    \label{fig:hese-vs-qt-comp}
\end{figure}

Figure~\ref{fig:hese-vs-qt-comp} shows the relative impact of TR and HESE in terms of classification accuracy by measuring them in isolation. The HESE and QT settings (without TR) apply term truncation by keeping the top $k$ terms in each individual weight. In this case, $\alpha$ is equal to $k$ as the group size is 1 (i.e., there is no grouping). We see that HESE substantially outperforms QT until the top $4$ terms are kept per weight ($\alpha = 4$) due to it requiring fewer terms. For the settings with TR, QT + TR and HESE + TR, we use a group size of $g=8$, with term budget $k$ values of $8$, $12$, $16$, $20$, and $24$ to generate comparable values of $\alpha$ as in the settings without TR. We find that TR improves the performance of both the QT and HESE encoding methods, with HESE + TR achieving the best performance.

\subsection{Quantization Error Analysis}
\label{sec:quant-error-analysis}

\begin{figure}
    \centering
    \includegraphics[width=\columnwidth]{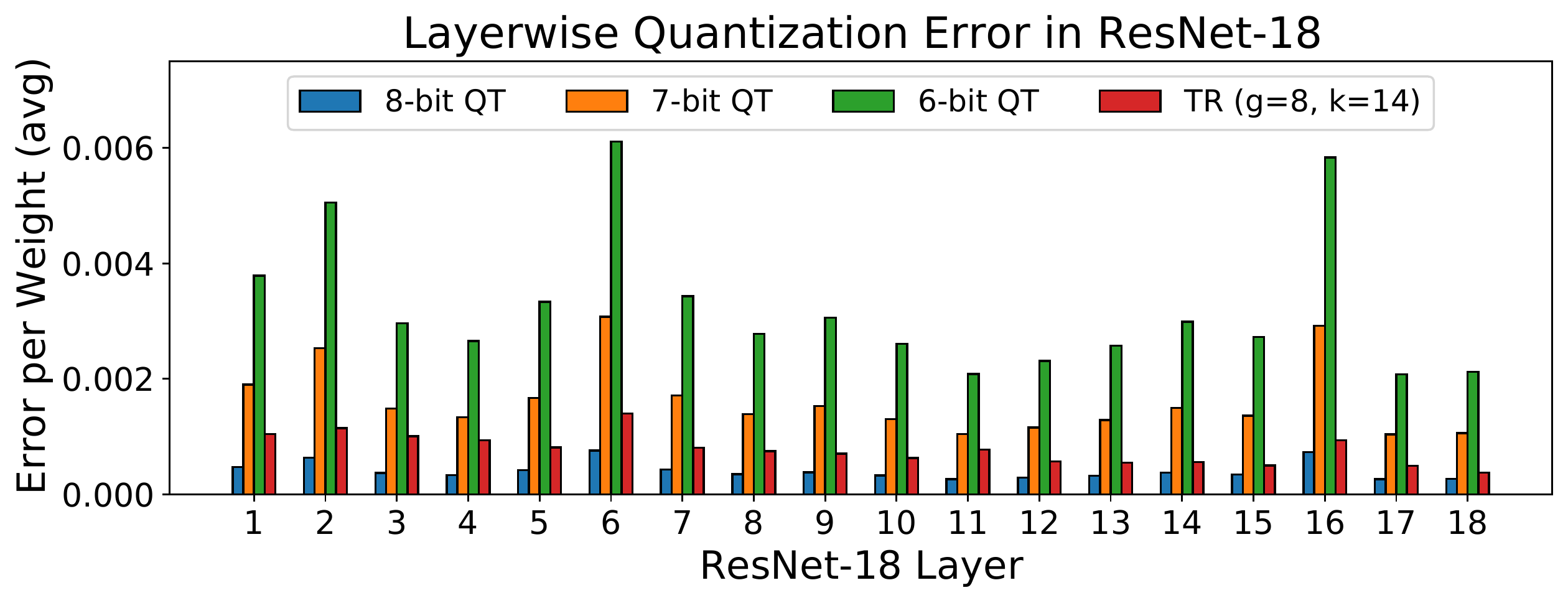}
    \caption{The average quantization error (relative to the original 32-bit floating-point weights) across the convolutional layers in ResNet-18  for 3 QT settings and one TR setting.}
    \label{fig:quant-error-analysis}
\end{figure}

The reason for TR's superior performance (e.g., accuracy or perplexity) over QT discussed in Section~\ref{sec:tr-vs-quant-acc} is due to TR introducing less quantization error. Figure~\ref{fig:quant-error-analysis} shows the quantization error across the layers in ResNet-18 for 3 QT settings (from 6-bit to 8-bit) and TR with a group size $g=8$ and a group budget $k=14$. We see that TR introduces a small amount of quantization error over 8-bit QT, which makes sense as TR is applied on top of 8-bit QT. The 7-bit and 6-bit QT settings truncate the low-order terms, leading to larger quantization error and reduced classification accuracy.

\section{FPGA Evaluation}
\label{sec:hw-eval}

In this section, we evaluate the hardware performance of the TR system described in Section~\ref{sec:term revealing-hw}. We have synthesized our TR system using Xilinx VC707 FPGA evaluation board. We first compare the performance of tMAC against a bit-parallel MAC in Section~\ref{sec:tmac-pmac-comparison}. Then, we demonstrate that our TR system can be used to implement both QT and TR in Section~\ref{sec:fpga-evaluation}. Finally, in Section~\ref{sec:fpga-system-evaluation}, we compare our TR system against the other FPGA-based CNN accelerators on ResNet-18.

\subsection{Comparing Performance of Bit-parallel MAC and tMAC}
\label{sec:tmac-pmac-comparison}
In this section, we evaluate the hardware performance of a single tMAC by comparing it against a bit-parallel MAC (pMAC) shown in Figure~\ref{fig:pmac-tmac-a}. For both designs, we perform a group of MAC computation: $y_{out} = \sum_{i=1}^{g}x_{i}w_{i} + y_{in}$, where $y_{in}$, $y_{out}$, $x_{i}$ and $w_{i}$ are 32-bit, 32-bit, 8-bit and 8-bit, respectively, and $g=8$ is the number of elements in the weight and data vectors (\ie~the group size in TR). In one cycle, the pMAC performs an 8-bit multiplication between $x_{i}$ and $w_{i}$ and a 32-bit accumulation between the result and $y_{in}$. Therefore, $y_{out}$ is generated in $g = 8$ cycles. By comparison, the tMAC takes a variable number of cycles to process each multiplication in the group, depending on the number of term pairs in the multiplication. In total, it requires no more than $s \times k$ cycles, where $s$ is the maximum number of terms for each data value and $k$ is the group term budget.

Table~\ref{table:resource-comparison} shows the FPGA resource consumption of the two MAC designs in terms of LookUp Tables (LUTs) and Flip-flops (FFs). The tMAC consume $6.5\times$ less LUTs and $6.0\times$ FFs than the pMAC. The tMAC requires less FPGA resources as it performs 3-bit exponent additions as opposed to 8-bit additions and 32-bit accumulation as in the pMAC.

We evaluate the two designs in terms of the energy efficiency, which is the ratio between the throughput and power consumption.  Table~\ref{table:energy-eff-comparison} shows the energy efficiency and classification accuracy for the two designs across four CNNs. For the tMAC settings, different values of $s$ and $k$ are selected for each CNN such that the classification accuracy stays competitive with the baseline model (less than $0.15\%$ difference in accuracy across all settings) while keeping the group size fixed ($g=8$). For each CNN, the energy efficiency of both MAC designs is normalized to that of the pMAC. We observe that tMAC achieves superior energy efficiency ($2.1\times$ on average) compared to pMAC across the four CNNs. This reflects that pMAC needs to perform more work than tMAC, as our analysis in Section~\ref{sec:tr-systolic} shows.

\begin{table}
\centering
\caption{FPGA resource consumption of pMAC and tMAC.}
\begin{adjustbox}{width=0.4\columnwidth,center}
\begin{tabular}{lcccc}
\hline
& LUT & FF \\  \hline
pMAC  &  154 &  148   \\
tMAC & 25 & 26     \\\hline

\hline
\end{tabular}
\end{adjustbox}
\label{table:resource-comparison}
\end{table}

\begin{table}
\centering
\caption{Classification accuracy and energy efficiency comparison for the two MAC designs across four CNNs.}
\begin{adjustbox}{width=\columnwidth,center}
\begin{tabular}{lcccccc}
\hline
Model                       & MAC  & $s$ & $k$ & $g$ & Accuracy & Energy Eff. \\ \hline
\multirow{2}{*}{Resnet-18}  & pMAC & -   & -   & -   & 69.62\%  & 1.0$\times$ \\
                            & tMAC & 3   & 12  & 8   & 69.60\%  & 2.1$\times$ \\ \hline
\multirow{2}{*}{VGG-16}     & pMAC & -   & -   & -   & 73.11\%  & 1.0$\times$ \\
                            & tMAC & 2   & 12  & 8   & 73.11\%  & 3.1$\times$ \\ \hline
\multirow{2}{*}{MoblNet-v2} & pMAC & -   & -   & -   & 71.76\%  & 1.0$\times$ \\
                            & tMAC & 3   & 18  & 8   & 71.65\%  & 1.5$\times$ \\ \hline
\multirow{2}{*}{EffNet-b0}  & pMAC & -   & -   & -   & 75.99\%  & 1.0$\times$ \\
                            & tMAC & 3   & 16  & 8   & 75.84\%  & 1.7$\times$ \\ \hline
\end{tabular}
\end{adjustbox}
\label{table:energy-eff-comparison}
\end{table}

\subsection{System Comparison of QT and TR}
\label{sec:fpga-evaluation}
In this section, we compare the hardware performance of TR against QT with the DNNs shown in Figure~\ref{fig:tr-quant-acc}. The systolic array in the TR system has 128 rows by 64 columns, with each systolic cell implementing a tMAC with a group size $g=8$. The group budget $k$ is chosen independently for each network such that the TR is within $0.15\%$ accuracy of the QT setting (or $0.05$ perplexity for the LSTM). In this section, we use the same TR system (Figure~\ref{fig:system}) for the implementation of both QT and TR in order to show the reconfigurability of our design. The implementation of QT does not require group-based ranking and HESE encoding, so we turn off these components of the hardware system to reduce dynamic power consumption. The control registers are configured based on Table~\ref{tab:hardware-components}.

\begin{figure}
  \centering
  \includegraphics[width=0.9\columnwidth]{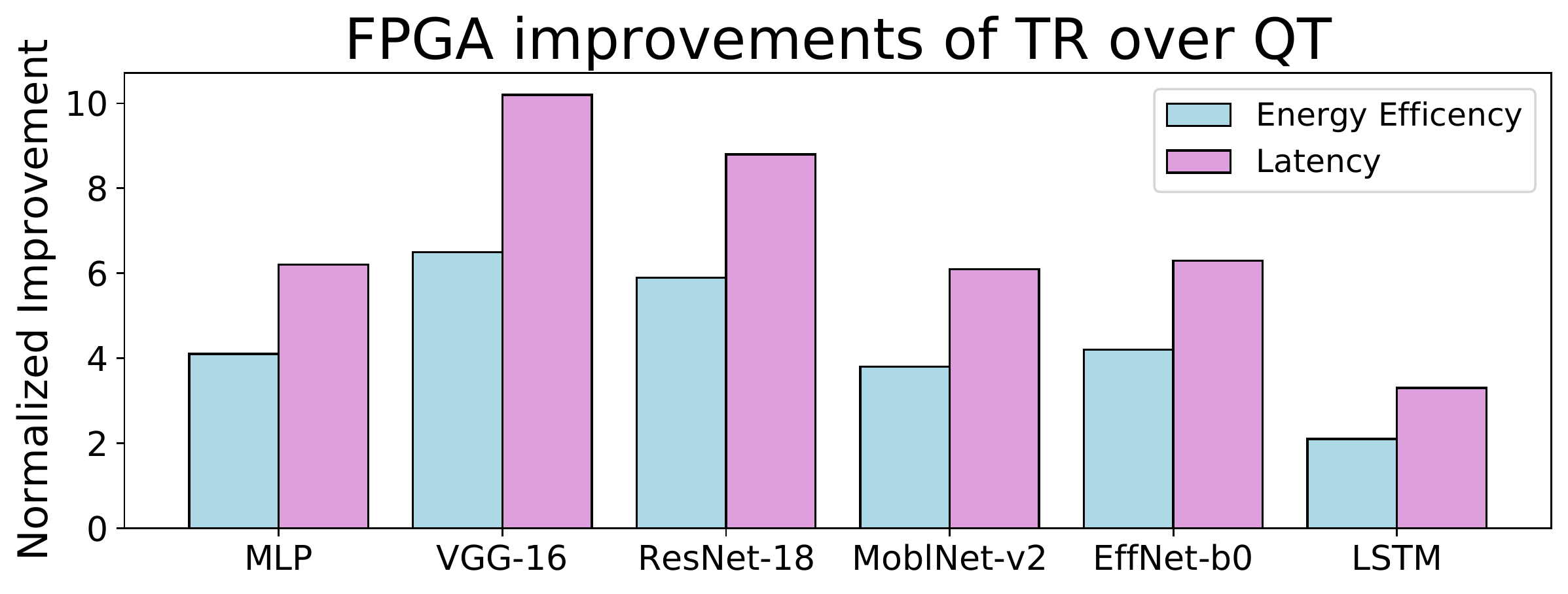}
  \caption{Normalized energy efficiency and latency improvements of TR over QT. All models use a group size of $g=8$. The group budget $k$ is selected for each model such that it is within 0.15\% accuracy of the corresponding QT setting ($k$ is $8$, $12$, $12$, $18$, $16$, $20$ for MLP, VGG-16, ResNet-18, MobileNet-V2, EfficientNet-b0, and LSTM, respectively). All models keep the top $s=3$ terms except for VGG-16 which uses $s=2$.}
\label{fig:fpga-models}
\end{figure}

We evaluate our FPGA system with the following performance metrics: (1) \textit{Average Processing latency} of the hardware system to generate the prediction result, and (2) \textit{Energy efficiency} or the average amount of energy required to process a single input sample. As shown in Figure~\ref{fig:fpga-models}, our TR system outperforms the QT by $7.8\times$ and $4.3\times$ on average in terms of processing latency and energy efficiency, respectively. For more difficult tasks, such as Wikitext-2 for the LSTM, a more conservative group budget $k=20$ is selected, leading to less relative improvement over QT. For overprovisioned models (e.g., VGG-16), a more aggressive group budget of $k=8$ is used, leading to more substantial improvements in latency and energy efficiency.

\subsection{FPGA System Evaluation}
\label{sec:fpga-system-evaluation}
In this section, we evaluate our TR system over ResNet-18 on ImageNet, using a group size $g=8$ and group budget $k=16$. While an even larger group size could theoretically lead to additional savings, there are diminishing returns as shown in Figure~\ref{fig:tr-group-size} when comparing $g=8$ and $g=32$. Additionally, larger group sizes increase the complexity of the term comparator due to additional tree levels of A\&C blocks.

We compare our TR system with the other FPGA-based accelerators which implements different CNN architectures (e.g., AlexNet) on ImageNet. We evaluate our design in terms of the average processing latency for the input samples, energy efficiency of the hardware system and classification accuracy. As shown in Table~\ref{table:fpga-evaluation}, our design achieves the highest classification accuracy $(69.48\%)$, energy efficiency ($25.22$ frames/J), and the second lowest latency ($7.21ms$). 

\begin{table}
\centering
\caption{Comparison of our FPGA implementation of ResNet-18 to other FPGA-based accelerators on ImageNet.}
 \begin{adjustbox}{width=\columnwidth,center}
\begin{tabular}{lcccccccc}
\hline
&\cite{zhang2018dnnbuilder} &\cite{shen2017maximizing} &\cite{qiu2016going}&\cite{xiao2017exploring}& Ours\\ \hline
FPGA Chip  & VC706 & Virtex-7 &  ZC706&  ZC706    &  VC707        \\
Acc. (\%) & 53.30\% & 55.70\%   & 64.64\%  & N/A      & 69.48\% \\
Frequency \small{(MHz)} & 200  & 100      & 150    & 100     &   170  \\
FF   & 51k(12\%) &348k(40\%)      & 127k(29\%)   & 96k(22\%)        & 316k(51\%)   \\
LUT  & 86k(39\%)  & 236k(55\%)     & 182k(83\%)   & 148k(68\%)     &201k(65\%) \\
DSP   & 808(90\%) &3177(88\%)    & 780(89\%) & 725(80\%)     & 756(27\%) \\
BRAM   & 303(56\%) & 1436(49\%)     & 486(86\%)  & 901(82\%)   & 606(59\%) \\
Latency (ms)  & 5.88  &11.7     & 224   & 17.3    & 7.21           \\
Energy eff. (frames/J)  & 23.6  &8.39   & 0.46   & 6.13   & 25.22          \\
\hline
\end{tabular}
\end{adjustbox}
\label{table:fpga-evaluation}
\end{table}

Our hardware system achieves the best performance for multiple reasons. First, TR coupled with the proposed HESE encoding greatly reduce the amount of term pair multiplications, which reduces the number of cycles in tMACs. TR allows tMACs to achieve a much tighter processing bound of $3 \times k$ pairs per group as opposed to $7 \times 7 \times g$ in the case of standard binary encoding without TR. Second, the bit-serial design of the coefficient accumulator in tMAC together with the systolic architecture of the computing engine leads to a highly regular layout with low routing complexity.

\section{Conclusion}
We proposed term revealing (TR) as a general run-time approach for furthering quantized computation on already quantized DNNs. Departing from conventional quantization that operates on individual values, TR is a group-based method that keeps a fixed number of terms within a group of values. TR leverages the weight and data distributions of DNNs, so that it can achieve good model performance even with a small group budget. We measure the computation cost of TR-enabled quantization using the number of term pair multiplications per inference sample. Under this clearly defined cost proxy, we have shown that TR significantly lowers computation costs for MLPs, CNNs, and LSTMs. As shown in Section~\ref{sec:fpga-evaluation}, this reduction in operations translates to improved energy efficiency and reduced latency over conventional quantization for our FPGA system. Furthermore, our FPGA system demonstrates that by changing a small number of control bits we can reconfigure a quantized computation under conventional quantization to one under TR-enabled quantization, and vice versa (Table~\ref{tab:hardware-components}). Quantization is one of most widely used approaches in streamlining DNNs; TR proposed in this paper brings the success of the quantization approach to another level.


\bibliographystyle{ieeetr}
\bibliography{references.bib}

\end{document}